%% file: paper.tex
\PassOptionsToPackage{dvipsnames}{xcolor}
\documentclass{gvc_research}
\usepackage[T1]{fontenc}
\usepackage{lmodern}
\usepackage{multirow}
\usepackage{booktabs}
\usepackage{soul}
\usepackage{pifont}
\usepackage[T1]{fontenc}
\usepackage{graphicx}
\usepackage{amsmath,amssymb}
\usepackage{tikz}
\usetikzlibrary{arrows.meta,positioning,calc,fit,shapes.geometric,decorations.pathreplacing}

\usepackage{makecell}
\usepackage{color, colortbl}
\usepackage{caption}
\usepackage{algorithm}

\input{preamble}

\usepackage{newtxtext}
\usepackage{enumitem}

\usepackage{algpseudocode}
\definecolor{catgray}{gray}{0.92}

\definecolor{FutureOrange}{HTML}{ff4a05}
\definecolor{teal}{HTML}{e37133}
\definecolor{orange}{HTML}{ff4a05}
\definecolor{purple}{HTML}{e37133}
\definecolor{blue}{HTML}{0d00b2}
\definecolor{myred}{HTML}{bf0511}
\definecolor{myblue}{HTML}{1272b6}
\newcommand{\SHORTNAME}{EditaLive}
\newcommand{\ours}{\textcolor{FutureOrange}{\textbf{EditaLive }}}
\newcommand{\KVSet}{\mathbf{KV}}
\newcommand{\ModelOutput}{\mathbf{X}_{\theta}}

\newcommand{\KVOutput}{\mathbf{kv}}
\newcommand{\optimal}[1]{\textcolor{myred}{\textbf{#1}}}
\newcommand{\suboptimal}[1]{\textcolor{myblue}{\textbf{#1}}}

\title{\textbf{EditaLive! Unified Character Video Editing for Live Streaming}}

\author[1,3]{Zhiyuan Li}
\author[1,\dagger]{Chi-Man Pun}
\author[2,\dagger]{Peng-Tao Jiang}
\author[2]{Bo Li}
\author[3,\P]{Xiaodong Cun}
\affiliation[1]{\textbf{University of Macau}}
\affiliation[2]{\textbf{vivo BlueImage Lab}}
\affiliation[3]{\textbf{GVC Lab, Great Bay University}}

\contribution[\dagger]{Corresponding Authors}
\contribution[\P]{Project Lead.}

\date{August 27th, 2026}

\adobedata[Project Page]{\url{https://huai-chang.github.io/EditaLive/}}

\begin{document}

\maketitle

\begin{center}
  \includegraphics[width=\linewidth]{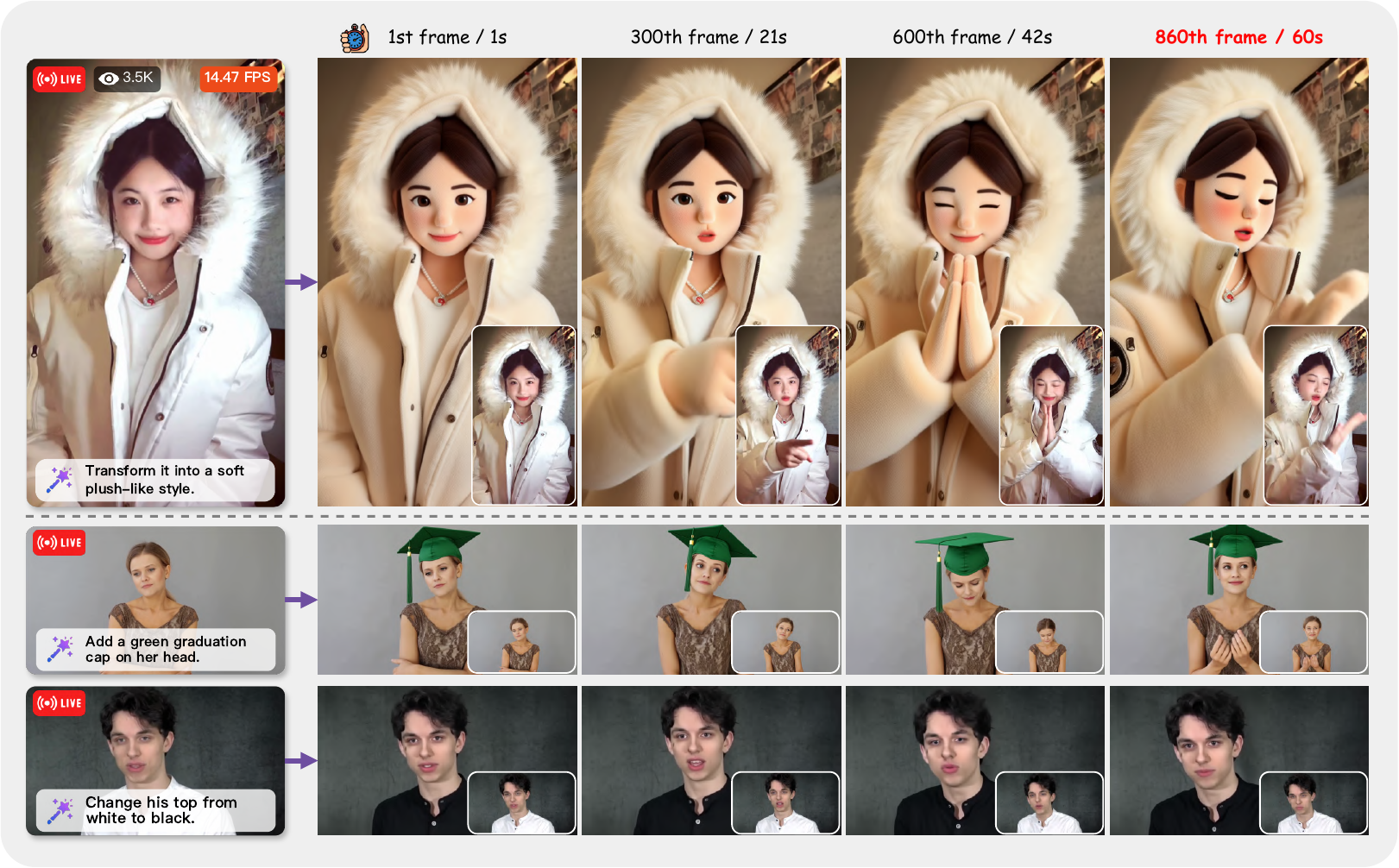}
  \captionof{figure}{An overview of the editing results and inference efficiency of \SHORTNAME. \SHORTNAME{} performs diverse local and global appearance edits in real time while faithfully preserving source motion and maintaining a consistent edited appearance over long sequences. Its appearance--motion decoupled formulation further supports cross-character editing.}
  \label{fig:teaser}
\end{center}

\abstract{
\input{sections/abstract}
}
\input{sections/1_introduction.tex}
\input{sections/2_relatedwork.tex}
\input{sections/5_data}

\input{sections/3_method}
\input{sections/6_experiments}
\input{sections/8_conclusion.tex}

\clearpage
\newpage
\bibliographystyle{assets/plainnat}
\bibliography{paper}

\input{sections/appendix}

\end{document}

%% file: preamble.tex
\usepackage{xspace}
\makeatletter
\DeclareRobustCommand\onedot{\futurelet\@let@token\@onedot}
\def\@onedot{\ifx\@let@token.\else.\null\fi\xspace}

\makeatother

%% file: sections/abstract.tex
Conventional video editing primarily focuses on scene-level content, whereas live streaming places greater emphasis on the human subject. 
However, directly applying existing video-editing methods to human-centric live streaming remains challenging, as they may introduce facial-expression inconsistencies and typically depend on multiple offline inference steps, making them unsuitable for real-time interaction. 
We propose \SHORTNAME, a novel framework for real-time streaming character video editing.
In detail, we start from a pretrained image animation model (Wan-Animate), which naturally decouples appearance from motion, and repurpose it as the base model for instruction-based human-centric video editing by reference frame editing and video reconstruction via the collected CharEdit-50K dataset.
Besides, we adapt the model from offline bidirectional to causal streaming generation, and design an aligned self-rollout distillation strategy that compresses the model into a two-step sampler, where fixed RoPE and align forcing reduce training--inference discrepancies, and first-frame preserved sparse attention filters redundant historical information to mitigate appearance drift.
Extensive experiments demonstrate that \SHORTNAME{} delivers state-of-the-art editing performance with faithful preservation of facial expressions and low-latency real-time streaming inference.

%% file: sections/1_introduction.tex
\section{Introduction}
\label{sec:introduction}

Live streaming has become one of the most popular forms of online entertainment, with streamers altering their on-screen looks to express their personalities and capture viewers' attention. In the era of generative AI, we define and tackle this task as \textit{streaming character video editing}, enabling streamers/viewers to customize live-streaming content using simple text prompts.

However, most of the current video editing methods are applied to the scene-level or object-level content. Early frame-by-frame editing approaches~\citep{live2diff, svdiff, streamdiffusion, streamv2v} apply image editing models to individual video frames, inevitably leading to temporal flickering. In contrast, recent native video editing algorithms~\citep{fatezero, lucyedit, ditto} perform joint video-level editing, substantially improving temporal consistency and editing quality, as shown in Fig.~\ref{fig:architecture}~(a). Despite their impressive performance, directly extending native video editing algorithms to character editing in live streaming scenarios is far from straightforward. This stems from three primary challenges:
\textit{(i) Facial expression mismatch}. Current video editing datasets~\citep{insvie, openve, ditto, ffp} are predominantly synthesized using video generation and editing models~\citep{vace, wan}, inevitably inheriting facial expression inconsistencies from these models, as shown in Fig.~\ref{fig:datapipeline}~(left). Consequently, models trained on such datasets struggle to faithfully preserve facial expressions.
\textit{(ii) Offline inference paradigm}. Existing instruction-based video editing models~\citep{univideo, editverse, lucyedit, ditto} typically require dozens of denoising steps together with classifier-free guidance (CFG)~\citep{cfg} to achieve high editing quality, while relying on bidirectional temporal attention over complete video clips to maintain temporal consistency. These offline-oriented designs make existing video editing models unsuitable for live streaming scenarios.
\textit{(iii) Long-term drift}. Although recent methods~\citep{liveedit, sana-streaming} adopt Self Forcing~\citep{self-forcing} for real-time video editing, their training and inference procedures remain misaligned in terms of RoPE indexing and KV-cache construction. The resulting errors accumulate during autoregressive inference, causing progressive appearance drift and undermining editing stability over long video sequences.

\input{figs/architecture}

For character video editing, we posit that character appearance and motion are naturally decoupled and can therefore be modeled separately. As the edited appearance remains consistent throughout the video stream, the task may not necessitate extensive denoising steps. Meanwhile, temporal consistency can be effectively maintained through causal temporal modeling over past contexts, without relying on offline bidirectional attention over complete video clips.

We thus propose \SHORTNAME, a unified framework for real-time streaming character video editing built on three key components. 
\textbf{(i) Appearance--Motion Decoupled Editing}. We build \SHORTNAME{} on Wan-Animate~\citep{wan-animate}, a pretrained image animation model that naturally decouples character appearance from motion conditions. We repurpose this model for instruction-based character video editing by combining reference-frame editing with video reconstruction, as shown in Fig.~\ref{fig:architecture}~(c). For each source video, we edit its reference frame with a forward instruction and construct a corresponding reverse instruction to recover the original appearance. Conditioned on the edited reference, reverse instruction, and source motion signals, \SHORTNAME{} is trained to reconstruct the source video, thereby learning diverse appearance transformations from paired source and edited appearances. Based on this formulation, we construct CharEdit-50K, a comprehensive dataset covering diverse character appearance editing tasks.
\textbf{(ii) Causal Streaming Adaptation}. Inspired by causal and autoregressive video generation methods~\citep{diffusionforcing, causvid}, we subsequently adapt the offline editing model to chunk-wise causal streaming generation, preserving bidirectional attention within each chunk while restricting cross-chunk attention to preceding context.
\textbf{(iii) Aligned Self-Rollout Distillation}. After adapting the model for causal streaming generation, we further employ self-rollout distillation~\citep{self-forcing} to compress it into a two-step sampler. Although self-rollout mitigates exposure bias~\citep{exposurebias2, exposurebias1} by training the model on its own autoregressive predictions, training--inference discrepancies in positional encoding and KV-cache construction can still lead to unstable long-term streaming generation. To address this, we introduce Fixed RoPE for consistent positional encoding and Align Forcing for inference-aligned KV-cache construction. Furthermore, we propose First-frame Preserved Sparse Attention~(FPSA) to filter out redundant historical information while keeping the first-frame features fully visible, effectively mitigating appearance drift during long-term generation.
Extensive quantitative and qualitative results show that \SHORTNAME{} achieves superior editing quality, expression consistency, and long-term stability, while enabling real-time streaming inference. Our contributions can be summarized as:
\begin{itemize}[nolistsep,leftmargin=*]
\setlength{\itemsep}{3pt}
\setlength{\parskip}{0pt}
\setlength{\parsep}{0pt}
\item We propose \SHORTNAME, a unified framework for real-time streaming character video editing that supports both diverse character appearance editing tasks and motion-driven animation.
\item We introduce an appearance--motion decoupled editing paradigm that reframes character video editing as an appearance transformation under explicit motion control, thereby preserving facial expressions. We further design a reconstruction-based training strategy built on motion-aligned real-video supervision and construct CharEdit-50K.
\item We develop an aligned self-rollout distillation strategy equipped with fixed RoPE, align forcing, and first-frame preserved sparse attention, effectively reducing training--inference discrepancies and mitigating appearance drift during long-term generation.
\end{itemize}

%% file: figs/architecture.tex
\begin{figure*}[ht]
    \centering
    \vspace{-0.5em}
    \includegraphics[width=\textwidth]{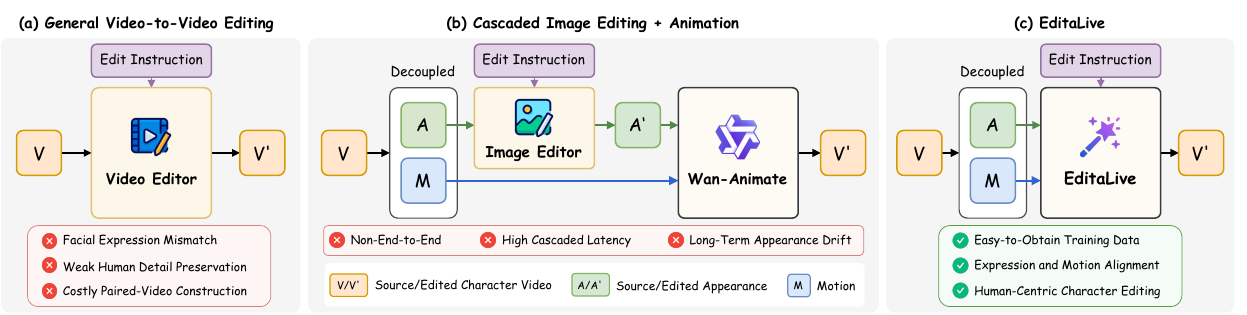}
    \vspace{-1.5em}
    \caption{
    Comparison of three character video editing paradigms.
    (a) General Video-to-Video Editing directly transforms a source video $V$ into an edited video $V'$, but often relies on costly synthetic video pairs that may contain facial-expression mismatches and weakened human details.
    (b) Cascaded image editing and animation first edits the source appearance $A$ into $A'$ and then animates it using motion $M$, introducing cascaded latency and potential long-term appearance drift.
    (c) \SHORTNAME{} unifies appearance editing and motion-conditioned generation, learning from edited reference images and motion-aligned real videos to preserve facial expressions and body motion while enabling stable, low-latency streaming editing.
    }
    \label{fig:architecture}
    \vspace{-0.5em}
\end{figure*}

%% file: sections/2_relatedwork.tex
\section{Related Work}
\label{sec:relatedwork}

\noindent\textbf{Instruction-based Image and Video Editing.}
Recent advances in text-to-image diffusion models~\citep{ldm, dit} have significantly improved instruction-based image editing, enabling users to modify visual content through natural language instructions~\citep{instructpix2pix, magicbrush, emu-edit, icedit}. 
Some works extend image editing to videos by applying image editing models to individual frames~\citep{live2diff, svdiff, streamdiffusion, streamv2v}. While efficient, these frame-wise methods lack stable temporal modeling and often lead to flickering and inconsistent edits.
Recent instruction-based video editing methods instead perform joint video-level editing with native video diffusion models~\citep{fatezero, vace, univideo, editverse, lucyedit, ditto}, achieving improved temporal consistency and editing quality. However, these methods are mainly designed for offline clip-level editing, relying on multi-step sampling and bidirectional temporal modeling over complete videos. They also commonly depend on synthetic editing datasets~\citep{insvie, ditto}, which can introduce facial expression inconsistencies in character-centric scenarios. 
More recent studies~\citep{liveedit, sana-streaming} adopt Self Forcing~\citep{self-forcing} to enable real-time causal streaming video editing. Nevertheless, they primarily focus on streaming efficiency and do not explicitly address stability over long generation horizons.
Unlike these methods, we focus on real-time streaming character video editing with explicit motion preservation and long-term generation capability.

\noindent\textbf{Motion-driven Character Animation.}
Motion-driven character animation aims to animate a source character according to motion signals extracted from a driving video. To achieve this, recent methods condition video generation on reference images and diverse motion representations, such as skeleton sequences~\citep{animateanyone, steadydancer}, 3D keypoints~\citep{scail}, motion frames~\citep{scail-2}, and implicit facial representations~\citep{x-nemo, wan-animate}. These works validate the effectiveness of separating appearance and motion for character generation. Building on this, recent efforts such as PersonaLive~\citep{personalive} further achieve real-time portrait animation through efficient distillation and micro-chunk streaming generation. However, most existing methods focus on animation or reenactment, where the source appearance is assumed to be fixed, rather than instruction-based character appearance editing. In contrast, we apply the idea of appearance--motion decoupling to character video editing, enabling appearance editing under explicit motion control while naturally supporting motion-driven animation by swapping character images.

\noindent\textbf{Efficient and Streaming Video Generation.}
Efficiency and streaming capability are critical for deploying video generation models in interactive and real-time scenarios. Existing methods commonly accelerate inference through model quantization~\citep{mpq-dm, sana}, sparse attention~\citep{vsa, liveditor}, and sampling-step distillation~\citep{dmd, dmd2}. To achieve real-time streaming video generation, recent works further introduce autoregressive paradigms~\citep{diffusionforcing, streamdiffusion}. CausVid~\citep{causvid} adapts diffusion video generation to causal inference. Self Forcing~\citep{self-forcing} trains the model with its own autoregressive predictions to reduce exposure bias. Building on these advances, we extend autoregressive streaming generation to character video editing and introduce an aligned self-rollout distillation strategy that enables efficient two-step inference while preserving character appearance over long generation horizons.

%% file: sections/5_data.tex
\section{Preliminaries}
\label{sec:preliminaries}

\noindent\textbf{Video Diffusion Models.} Video diffusion models learn a conditional distribution over video sequences through iterative denoising. To reduce the computational cost of processing high-dimensional videos, an input video $x$ is typically encoded into a compact latent representation $Z_1$ using a spatio-temporal VAE, and generation is performed in the resulting latent space. Under the flow-matching formulation~\citep{fm}, a noisy latent $Z_t$ is constructed by interpolating between a Gaussian sample $Z_0\sim\mathcal{N}(0,I)$ and the clean video latent:
\begin{equation}
    Z_t = tZ_1 + (1-t)Z_0,\qquad t\in[0,1].
\end{equation}
The corresponding target velocity $V_t=\frac{\mathrm{d}Z_t}{\mathrm{d}t}=Z_1-Z_0$ is constant along this path. A conditional velocity model $v_\theta$ is trained to estimate $V_t$ from the interpolated latent $Z_t$, timestep $t$, and conditioning signals $c$ by minimizing
\begin{equation}
    \mathcal{L}_{\mathrm{FM}}
    =
    \mathbb{E}_{Z_1,Z_0,t}
    \left[
        \left\|
            v_\theta(Z_t,t\mid c)-(Z_1-Z_0)
        \right\|_2^2
    \right],
\end{equation}
where $c$ may include text, image, or other conditions. At inference time, a video latent is generated by numerically integrating the learned velocity field from the Gaussian prior at $t=0$ toward the data distribution at $t=1$, after which the VAE decoder maps the resulting latent back to video space.

\noindent\textbf{Distribution Matching Distillation.} 
Distribution Matching Distillation (DMD)~\citep{dmd, dmd2} distills a pretrained diffusion model into a student generator $G_\theta$ with substantially fewer denoising steps. Let $p_{\theta,t}$ denote the distribution obtained by perturbing student-generated samples to timestep $t$, and let $p_{\mathrm{real},t}$ denote the corresponding distribution induced by the real data or pretrained teacher model. DMD minimizes the reverse Kullback--Leibler divergence between these distributions:
\begin{equation}
    \mathcal{L}_{\mathrm{DMD}}
    =
    \mathbb{E}_{t}
    \left[
        D_{\mathrm{KL}}
        \left(
            p_{\theta,t}\,\|\,p_{\mathrm{real},t}
        \right)
    \right].
    \label{eq:dmd_objective}
\end{equation}
Given a clean prediction $\hat{Z}_1$ and its perturbed version $\hat{Z}_t$, the gradient of this objective can be estimated from the difference between the score functions of the student and target distributions:
\begin{equation}
    \nabla_\theta\mathcal{L}_{\mathrm{DMD}}
    =
    - \mathbb{E}_{Z_0,t}
    \left[
        \left(
            s_{\mathrm{real}}(\hat{Z}_t,t,c)
            -
            s_{\mathrm{fake},\phi}(\hat{Z}_t,t,c)
        \right)^{\!\top}
        \frac{\partial G_\theta(Z_0,c)}{\partial\theta}
    \right],
    \label{eq:dmd_gradient}
\end{equation}
where $s_{\mathrm{real}}$ estimates the score of the target distribution and $s_{\mathrm{fake},\phi}$ estimates the score of the current student distribution. The real-score model is kept fixed, whereas the fake-score model is trained on samples produced by the evolving student generator. Training alternates between updating $s_{\mathrm{fake},\phi}$ to track the student distribution and optimizing $G_\theta$ using the resulting score difference, progressively moving the few-step student generator toward the target distribution.

%% file: sections/3_method.tex
\section{\ours :\ Human-Centric Video Editing Model}
\label{sec:method}

In this paper, we define a \textit{character video} as a video centered on a primary human or human-like subject. Streaming character video editing aims to modify the appearance of the primary character in real time according to a user instruction. Formally, given a source video stream $x^{1:N}$ and an editing instruction $p$, the goal is to generate an edited stream $y^{1:N}$:
\begin{equation}
    y^{i} = \mathcal{F}(x^{i}, p, y^{< i}), i=1,2,\dots,N,
    \label{eq:task_define}
\end{equation}
where $\mathcal{F}$ denotes the editing model and $y^{<i}$ represents previously generated outputs. As illustrated in Fig.~\ref{fig:overview}, \SHORTNAME{} addresses
streaming character video editing through three consecutive stages. 
First, we decompose the source character video into appearance and motion representations, reformulating character video editing as appearance editing conditioned on the explicit motion signals (Sec.~\ref{sec:decouple}). 
Second, we adapt the offline bidirectional video model to causal streaming generation, enabling continuous inference based on cached historical context (Sec.~\ref{sec:temporal}). 
Finally, we develop an aligned self-rollout distillation strategy that compresses the causal model into a two-step sampler while reducing training--inference discrepancies and improving long-term stability (Sec.~\ref{sec:align}). 
Together, these components form a unified framework for efficient streaming character editing with explicit motion preservation and stable long-term generation.

\input{figs/overview}

\subsection{Appearance--Motion Decoupled Editing}
\label{sec:decouple}
As shown in Fig.~\ref{fig:overview}~(a), we reformulate character video editing by decoupling the appearance to be edited from the motion to be preserved. Given a source character video $x^{1:N}$, we select a reference frame $I$ to represent the character appearance, while extracting a skeleton sequence $s^{1:N}$ and implicit facial representations $f^{1:N}$ to capture the source body motion and facial dynamics, respectively. The editing process is then formulated as
\begin{equation}
    y^{1:N} = \mathcal{F}(I,p,s^{1:N},f^{1:N}),
    \label{eq:decoupled_editing}
\end{equation}
where the reference frame $I$ and instruction $p$ jointly determine the edited appearance.

\noindent\textbf{Reconstruction-based Training.} Synthetic source--target video pairs often exhibit facial expression mismatches (Fig.~\ref{fig:datapipeline}~(left)), entangling the desired appearance edit with unintended changes in facial dynamics. We therefore restrict synthetic editing to the reference image while retaining the real video as the reconstruction target. Specifically, given the reference image $I$, we first apply a forward instruction $p^{\mathrm{fwd}}$ via an off-the-shelf image editing model $\mathcal{E}_{\mathrm{img}}$ to obtain an edited image:
\begin{equation}
    \widetilde{I} = \mathcal{E}_{\mathrm{img}}(I, p^{\mathrm{fwd}}).
    \label{eq:forward_image_edit}
\end{equation}
We then construct a reverse instruction $p^{\mathrm{rev}}$ for restoring the original appearance from $\widetilde{I}$. Next, we encode the edited image $\widetilde{I}$ and the original video $x^{1:N}$ into the VAE latent space, obtaining the reference latent $z^{\mathrm{ref}}$ and target video latents $z^{\mathrm{tgt}}$, respectively. We define $Z_1 = (z^{\mathrm{ref}}, z^{\mathrm{tgt}})$ as the clean reference--target sequence and sample $Z_0\sim\mathcal{N}(0,\mathbf{I})$. At timestep $t\in[0,1]$, we construct
\begin{equation}
    Z_t = tZ_1 + (1-t)Z_0 = (z_t^{\mathrm{ref}},z_t^{\mathrm{tgt}}),
    \label{eq:reconstruction_flow_path}
\end{equation}
where $z_t^{\mathrm{ref}}$ and $z_t^{\mathrm{tgt}}$ denote the noised reference and target video latents, respectively. We then construct $H_t$ by channel-wise concatenating the noised reference–target sequence, the clean reference condition, and a reference mask:
\begin{equation}
    H_t = (z_t^{\mathrm{ref}},z_t^{\mathrm{tgt}}) \mathbin{\|} 
    (z^{\mathrm{ref}},\mathbf{0},\ldots,\mathbf{0}) \mathbin{\|} 
    (\mathbf{1},\mathbf{0},\ldots,\mathbf{0}),
    \label{eq:reconstruction_input}
\end{equation}
where $(\cdot,\cdot)$ and $\mathbin{\|}$ denote temporal and channel-wise concatenation, respectively. The model $v_{\theta}$ predicts the ground-truth flow velocity $V_t=\frac{\mathrm{d}Z_t}{\mathrm{d}t}=Z_1-Z_0$ and is optimized by
\begin{equation}
    \mathcal{L}_{\mathrm{FM}} = \mathbb{E}_{Z_1,Z_0,t}
    \left[ \left\| v_{\theta} \left( H_t,t \mid p^{\mathrm{rev}},
    s^{1:N}, f^{1:N} \right) - (Z_1-Z_0) \right\|_2^2\right].
    \label{eq:reconstruction_flow_matching}
\end{equation}
By treating the original video latent $z^{\mathrm{tgt}}$ as the reconstruction target, $\mathcal{L}_{\mathrm{FM}}$ forces the model to follow $p^{\mathrm{rev}}$ and recover the original appearance from the edited reference $\widetilde{I}$, while retaining the authentic facial dynamics and body motion.

\input{figs/datapipeline}

\noindent \textbf{CharEdit-50K.} Following the reconstruction-based formulation, we construct CharEdit-50K through the three-stage pipeline shown in Fig.~\ref{fig:datapipeline}~(right). First, we mine high-quality reference frames from a large collection of character videos. Second, GPT-5.5~\citep{gpt} generates candidate bidirectional instruction pairs for each reference image. Third, Qwen-Image-Edit~\citep{qwen} and Nano Banana 2~\citep{gemini} synthesize the corresponding edited images, which are subsequently filtered by GPT-5.5. Each retained sample contains an edited image, its reverse instruction, and the corresponding source video. Further dataset details are provided in Appendix~\ref{app:data}.

\subsection{Causal Streaming Adaptation}
\label{sec:temporal}

While bidirectional self-attention in the previous stage enables temporally consistent editing, it requires complete video clips as input and incurs computational and memory costs that grow rapidly with video length, making it unsuitable for streaming inference. We therefore adapt the model to chunk-wise causal generation, where frames interact bidirectionally within each chunk while different chunks are temporally ordered, as shown in Fig.~\ref{fig:overview}~(b). Let $z^{\mathrm{tgt}} = (z^{(1)},\ldots,z^{(K)})$ denote the target video latents divided into $K$ chunks. At timestep $t$, we perturb the target video latents to obtain $z_t^{\mathrm{tgt}}$, while keeping the reference latent $z^{\mathrm{ref}}$ clean. The input is constructed as
\begin{equation}
    H_t^{\mathrm{causal}} = (z^{\mathrm{ref}},z_t^{\mathrm{tgt}},z^{\mathrm{context}}) \mathbin{\|}
    (z^{\mathrm{ref}},\mathbf{0},\ldots,\mathbf{0}) \mathbin{\|}
    (\mathbf{1},\mathbf{0},\ldots,\mathbf{0}),
    \label{eq:causal_input}
\end{equation}
where $z^{\mathrm{context}} = (z^{(1)},\ldots,z^{(K-1)})$ denotes the clean context. Unlike stage~1 in Sec.~\ref{sec:decouple}, the reference latent in the first term is directly used as clean context rather than being perturbed. We apply a chunk-wise causal attention mask such that each noisy target chunk attends to the clean reference latent, its own latents, and the clean context chunks preceding it, but not to other noisy target chunks or future context. This design enables all target chunks to be trained in parallel while preserving the causal dependencies required for streaming inference. The Flow Matching~\citep{fm} loss is computed over the positions corresponding to \(z_t^{\mathrm{tgt}}\), while the reference and clean-context positions are excluded from the loss and used solely as conditioning.

\subsection{Aligned Self-Rollout Distillation}
\label{sec:align}

Although causal adaptation enables streaming inference, the model still requires multi-step sampling and suffers from error accumulation during long-term generation. Building on Self Forcing~\citep{self-forcing}, we propose an aligned self-rollout distillation strategy that reduces the sampling process to two steps and matches rollout training with streaming inference, as shown in Fig.~\ref{fig:overview}~(c). Specifically, Align Forcing eliminates discrepancies in KV-cache construction, while Fixed RoPE maintains consistent positional encoding between training and inference. We further introduce First-frame Preserved Sparse Attention to filter redundant historical context and mitigate appearance drift.

\input{figs/alignforcing}

\noindent \textbf{Align Forcing.} As shown in Fig.~\ref{fig:alignforcing}~(a), Self Forcing~\citep{self-forcing} couples the randomly selected gradient-update step with rollout termination. Once the selected step is reached, the rollout stops and the clean estimate at that step is used to construct the KV cache propagated to subsequent chunks. During streaming inference, however, each chunk completes the entire denoising trajectory, and only the final prediction is used to construct the propagated state. This creates a training--inference mismatch in inter-chunk state construction: training may propagate an intermediate rollout state, whereas inference always propagates the final state. To eliminate this discrepancy, we decouple gradient update from rollout termination. As shown in Fig.~\ref{fig:alignforcing}~(b), gradients are computed only at the selected step, while the remaining denoising steps are completed with stop-gradient. The KV cache is then constructed from the final prediction and propagated to subsequent chunks, always matching the inference stage (Fig.~\ref{fig:alignforcing}~(c)). Under two-step distillation, Self Forcing requires an average of $2.5$ NFEs per training iteration, whereas Align Forcing requires 2 NFEs, with only the first chunk incurring one additional $t_0$ cache forward to initialize the attention sink.

\noindent \textbf{Fixed RoPE.} Standard RoPE assigns temporal indices according to the absolute latent positions. Since self-rollout training covers only a finite number of chunks, long-term inference eventually encounters indices outside the training range. Meanwhile, the reference sink remains at position $0$, causing its positional offsets from later latents to grow continuously and progressively weakening appearance conditioning. We therefore introduce Fixed RoPE, which applies the same positional layout $0$--$9$ during both rollout training and streaming inference, as shown in Fig.~\ref{fig:overview}~(c). Specifically, the reference sink, attention sink, local-window context, and current chunk are assigned indices $0$, $1$--$3$, $4$--$6$, and $7$--$9$, respectively. By assigning positions according to their roles in the streaming context rather than their absolute timestamps, Fixed RoPE avoids positional extrapolation and keeps the appearance reference at a fixed distance from the latents currently being generated.

\noindent \textbf{First-frame Preserved Sparse Attention.}
Since the local-window context is temporally and visually closest to the current chunk, the current queries tend to over-rely on it while underutilizing the stable appearance cues in the reference and attention sinks. Inspired by VSA~\citep{vsa}, we partition the current query and cached key tokens into spatio-temporal blocks, and compute block-level query--key scores. For each query block, we retain the top-$K$ key blocks, where $K=(1-\rho)B$, $B$ denotes the total number of candidate key blocks and $\rho$ represents the sparsity ratio. Token-level attention over the original queries, keys, and values is then computed only within the selected block pairs. Since this content-adaptive selection may still favor the local window, we always include all key blocks from the first generated frame in the selected $K$ blocks and choose the remaining blocks according to their scores, as shown in Fig.~\ref{fig:overview}~(c). As a stable realization of the target appearance, the first frame serves as a persistent appearance anchor, thereby mitigating appearance drift during long-term generation.

%% file: figs/overview.tex
\begin{figure*}[t]
    \centering
    \includegraphics[width=\textwidth]{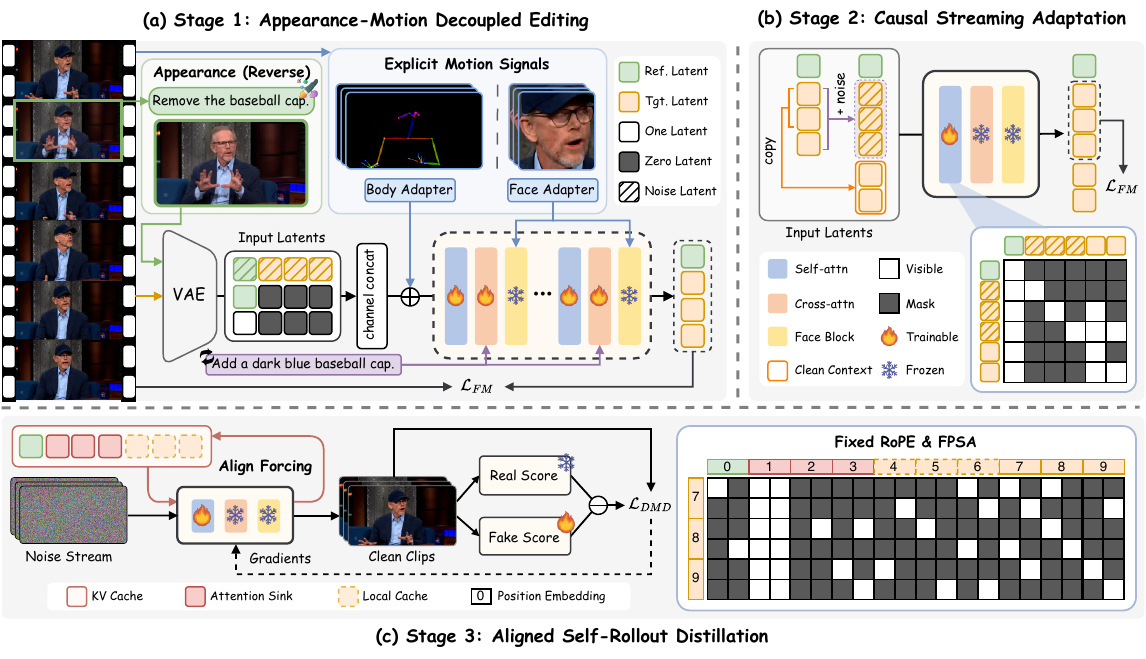}
    \vspace{-1.5em}
    \caption{Overview of the three-stage training pipeline of \SHORTNAME. (a) Appearance--Motion Decoupled Editing reformulates character video editing as appearance editing with explicit body and facial motion preservation. (b) Causal Streaming Adaptation converts bidirectional modeling into causal generation using clean historical context and causal attention. (c) Aligned Self-Rollout Distillation compresses the model into a two-step sampler, where Align Forcing and Fixed RoPE align training with streaming inference, and FPSA improves long-term appearance stability.}
    \vspace{-1em}
    \label{fig:overview}
\end{figure*}

%% file: figs/datapipeline.tex
\begin{figure*}[t]
    \centering
    \includegraphics[width=\textwidth]{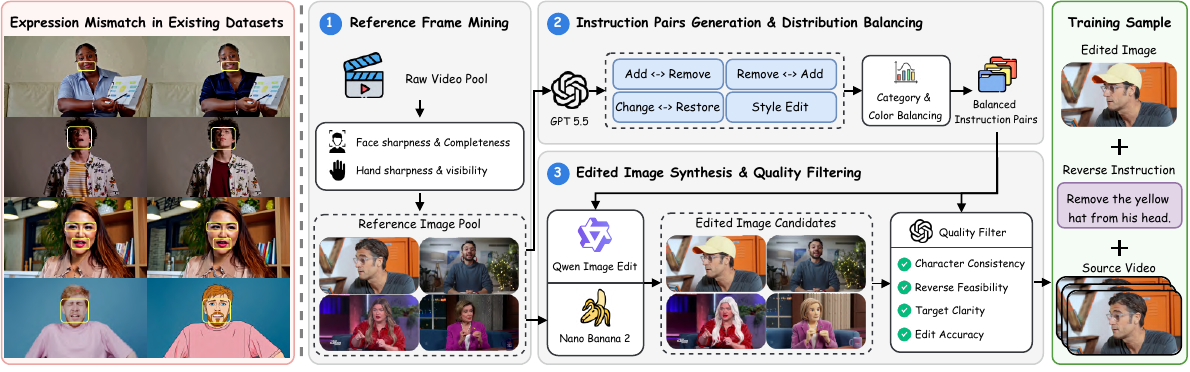}
    \vspace{-1.5em}
    % \caption{Construction pipeline of CharEdit-50K. (Left) Existing video editing datasets often contain facial expression inconsistencies. (Right) Our three-stage data pipeline.}
    % first mines high-quality reference frames, then generates and balances bidirectional instruction pairs, and finally synthesizes appearance-edited images followed by multidimensional quality filtering. Each dataset sample contains an edited image, its reverse instruction, and the corresponding authentic source video.}
    \caption{
    Construction pipeline of CharEdit-50K.
    (Left) Existing video editing datasets often contain facial expression inconsistencies.
    (Right) Our pipeline restricts synthetic editing to the reference image while retaining its authentic source video as motion-aligned supervision.
    (1) High-quality reference frames are selected based on face and hand quality.
    (2) GPT-5.5 generates instructions covering add--remove, remove--add, change--restore, and stylization, followed by category and color balancing.
    (3) Qwen-Image-Edit and Nano Banana 2 synthesize candidate images, which are filtered for edit accuracy, target clarity, character consistency, and reverse feasibility.
    Each retained bidirectional sample contains an edited image, its reverse instruction, and the corresponding source video.
    }
    \label{fig:datapipeline}
    \vspace{-1em}
\end{figure*}

%% file: figs/alignforcing.tex
\begin{figure*}[t]
    \centering
    \includegraphics[width=\textwidth]{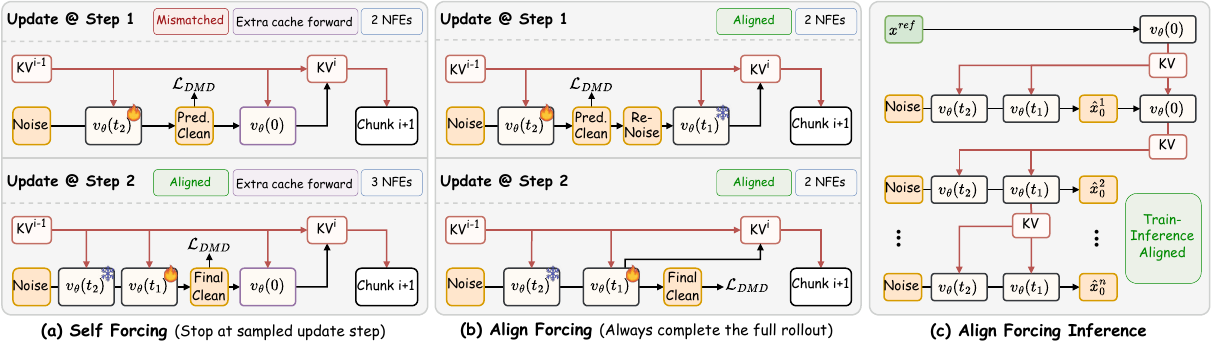}
    \vspace{-1.5em}
    % \caption{Comparison of Self-Forcing and Align Forcing under two-step distillation.}
    % (a) Self-Forcing may propagate an intermediate rollout state and requires $2.5$ NFEs on average. (b) Align Forcing always completes the full rollout before propagating the KV cache. (c) The corresponding two-step streaming inference process.}
    \caption{
    Comparison of Self Forcing and Align Forcing under two-step distillation.
    (a) Self Forcing couples gradient update with rollout termination, causing an intermediate prediction to be cached when the rollout stops early. This creates a mismatch with inference, which always caches the final prediction, and requires $2.5$ NFEs on average.
    (b) Align Forcing applies gradients only at the sampled step but completes the remaining rollout with stop-gradient. The KV cache is therefore always constructed from the final prediction, matching inference while requiring two NFEs.
    (c) During streaming inference, each chunk propagates its final KV cache after two-step denoising, with an additional $t_0$ forward pass for the first chunk to initialize the attention sink.
    }
    \vspace{-1em}
    \label{fig:alignforcing}
\end{figure*}

%% file: sections/6_experiments.tex
\section{Experiments}
\label{sec:experiments}

\subsection{Implementation Details}
We build our model on Wan-Animate~\citep{wan-animate} and train all three stages on CharEdit-50K using LoRA, with both the rank $r$ and scaling factor $\alpha$ set to $128$. Experiments are conducted on 8 NVIDIA H100 GPUs. All stages use the AdamW optimizer with a global batch size of $8$. 
Stages~1 and~2 are each trained on 33-frame, 480p video clips for 15K steps with a learning rate of $1\times10^{-4}$, while Stage~3 is trained on 69-frame, 480p clips for 2K steps with a learning rate of $2\times10^{-6}$. The Stage~3 student is initialized from the Stage~2 checkpoint, whereas its real-score and fake-score branches are initialized from the Stage~1 checkpoint. For Stages~2 and~3, we set the chunk size to $3$. During self-rollout distillation, the first generated chunk is retained as an attention sink. We set the sparsity ratio $\rho$ in FPSA to 75\%. For two-step generation, the sampling timesteps are set to $[1000, 250]$. We additionally construct CharEdit-Bench for comprehensive evaluation, comprising a short-video subset (CharEdit-Bench-S) and a long-video subset (CharEdit-Bench-L). Complete implementation details and benchmark specifics are provided in Appendix~\ref{app:implementation} and~\ref{app:bench}.

\subsection{Comparison with Other Methods}

\input{figs/qualitative}

We compare our method against state-of-the-art instruction-based video editing approaches, including the bidirectional baselines LucyEdit~\citep{lucyedit}, UniVideo~\citep{univideo}, and Ditto~\citep{ditto}, as well as the streaming baselines LiveEdit~\citep{liveedit}, SANA-Streaming~\citep{sana-streaming}, and StreamDiffusionV2~\citep{streamdiffusionv2}. We additionally include a cascaded image-editing-and-animation baseline that first edits the reference image using Qwen-Image-Edit~\citep{qwen} and then animates the edited image with Wan-Animate~\citep{wan-animate}. We evaluate the performance in terms of character consistency, editing quality, video quality, and inference efficiency. ID-SIM~\citep{arcface} measures identity consistency, while AED and APD~\citep{fomm} assess the preservation of facial expressions and body poses, respectively. A VLM-based evaluator~\citep{editverse} assesses overall editing quality. Pick Score~\citep{pickscore} is used to assess the overall quality of the edited videos. For efficiency, we report end-to-end generation throughput in frames per second (FPS) and average inter-chunk latency. Detailed baseline configurations, metric definitions, and evaluation protocols are provided in Appendix~\ref{app:baseline} and~\ref{app:evaluation}.

\noindent\textbf{Qualitative \& Quantitative Comparison.} As shown in Fig.~\ref{fig:qualitative}, \SHORTNAME{} consistently performs both local appearance edits and global style transformations while faithfully preserving the source facial expressions, body poses, and scene structure. In comparison, existing streaming methods may either fail to fully apply the requested edits or introduce noticeable inconsistencies in facial expression. The bidirectional baselines exhibit similar limitations, despite their substantially higher inference costs. These qualitative observations are consistent with the quantitative results in Table~\ref{tab:quant_rec}, where \SHORTNAME{} ranks first or second across all eight quality metrics. We omit the character-consistency metrics of LiveEdit and LucyEdit because they do not support global style editing, making these metrics incomparable when the requested transformation is not successfully applied. Additional comparison results are provided in Appendix~\ref{more_experiments}. 

\noindent \textbf{Efficiency.} As reported in Table~\ref{tab:quant_rec}, \SHORTNAME{} achieves an end-to-end generation throughput of $14.47$ FPS with an average inter-chunk latency of $0.829$ seconds on a single NVIDIA H100 GPU. Although the dedicated streaming baselines achieve higher raw efficiency, their edit success rates are at most $0.428$, substantially below the $0.720$ achieved by \SHORTNAME{}. Compared with LucyEdit, the fastest bidirectional baseline, \SHORTNAME{} delivers $4.7\times$ higher throughput and reduces latency by a factor of $31.7$. For the cascaded Qwen-Image-Edit and Wan-Animate baseline, even when excluding the one-time cost of reference-image editing, \SHORTNAME{} remains $11.4\times$ faster and reduces latency by a factor of $76.7$. These results demonstrate that \SHORTNAME{} provides a favorable quality--efficiency trade-off, combining sub-second streaming latency with substantially stronger editing performance. Moreover, replacing the standard Wan-VAE decoder with Flash-VAED~\citep{flash-vaed} further increases the inference speed of \SHORTNAME{} to 16.4 FPS.

\input{tables/comparison}

\input{figs/ablation}

\subsection{Ablation Studies}
To validate the effectiveness of our key components, we conduct comprehensive ablation studies on CharEdit-Bench-L, covering both causal streaming adaptation and aligned self-rollout distillation.

\noindent\textbf{Causal Streaming Adaptation.}
We first examine how the reference latent should be used during causal adaptation. As shown in Fig.~\ref{fig:ablation}~(left), optimizing both the reference and target positions (\textit{Ref.+Tgt.}) introduces conflicting supervision. Under the same instruction, the reference latent is optimized to reconstruct its unchanged input appearance, whereas the target latents are required to perform the instructed appearance transformation. This conflict weakens the reference conditioning and causes unintended changes to the generated character. We therefore keep the reference latent clean and use it solely as conditioning context, while computing the Flow Matching loss only over the noisy target positions (\textit{Tgt. only}). This target-only objective separates the clean reference condition from the optimization target, avoiding contradictory gradients and enabling faithful instruction-guided appearance transformation.

\noindent\textbf{Aligned Self-Rollout Distillation.}
We evaluate the key components of aligned self-rollout distillation in Table~\ref{tab:quant_abl} and Fig.~\ref{fig:ablation}~(right). 
Replacing Align Forcing with Self Forcing (\textit{w/o Align Forcing}) introduces a mismatch in KV-cache construction between rollout training and streaming inference, causing the largest performance degradation: ID-SIM decreases from $0.492$ to $0.252$, and SR decreases from $0.817$ to $0.367$.
Removing Fixed RoPE exposes the model to unseen positional offsets during long-video inference and weakens its reference conditioning, resulting in progressive appearance drift and an ID-SIM of only $0.381$.
FPSA contributes through both sparse context selection and explicit first-frame preservation. Without FPSA, full attention overemphasizes recent local context while underutilizing stable appearance cues, resulting in an ID-SIM of $0.455$. Introducing sparse attention without first-frame preservation (\textit{w/o FP}) improves ID-SIM to $0.466$, demonstrating the benefit of filtering redundant historical information. Nevertheless, visible appearance changes still emerge over time, such as the highlighted hair variation at frame $900$. Explicitly preserving the first-frame features further increases ID-SIM to $0.492$ and reduces AED and APD to $0.576$ and $0.109$, respectively.
Overall, the complete model achieves the best ID-SIM, AED, APD, TA, EQ, and BC, confirming that the three components jointly improve character consistency without sacrificing video quality or temporal coherence.

\input{tables/ablation}

%% file: figs/qualitative.tex
\begin{figure*}[t]
    \centering
    \includegraphics[width=\textwidth]{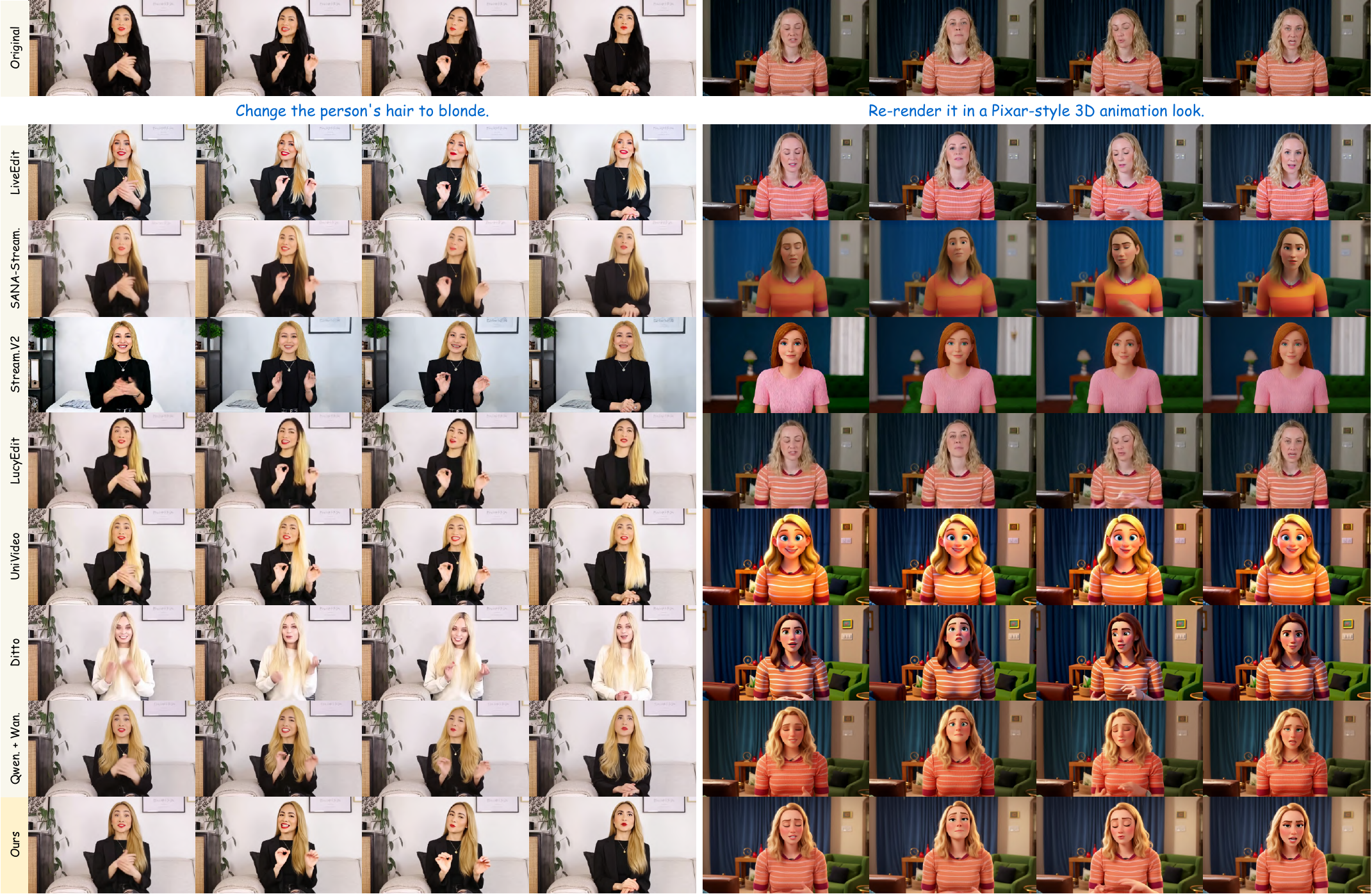}
    \vspace{-1.5em}
    \caption{Qualitative comparisons. \SHORTNAME{} faithfully follows the editing instructions while preserving the source facial expressions, body poses, and scene structure across frames.
    }
    \label{fig:qualitative}
    \vspace{-1em}
\end{figure*}

%% file: tables/comparison.tex
\begin{table*}[t]
\centering
\caption{Quantitative comparisons on CharEdit-Bench-S. Numbers in \textcolor{myred}{\textbf{red}} and \textcolor{myblue}{\textbf{blue}} indicate the best and the second-best results, respectively. APD multiplied by $10$. TA, EQ, BC and SR denote text alignment, edit quality, background consistency, and success rate, respectively. 
\textit{*LiveEdit and LucyEdit do not support global style editing. We therefore omit their character-consistency metrics, as these scores are not comparable when the requested edit is not successfully performed.}}
\label{tab:quant_rec}
\resizebox{\textwidth}{!}{
\begin{tabular}{lcccc|cccc|c|cc}
\toprule
\multirow{2}{*}{Method} &\multirow{2}{*}{\textbf{\#Params}} & \multicolumn{3}{c}{\textbf{Character Consistency}} &\multicolumn{4}{c}{\textbf{VLM Evaluation}} &\multicolumn{1}{c}{\textbf{Video Quality}} &\multicolumn{2}{c}{\textbf{Efficiency}}\\
\cmidrule(lr){3-5} \cmidrule(lr){6-9} \cmidrule(lr){10-10} \cmidrule(lr){11-12}
& & \textbf{ID-SIM}\ $\uparrow$ & \textbf{AED}\ $\downarrow$ & \textbf{APD}\ $\downarrow$ & \textbf{TA}\ $\uparrow$ & \textbf{EQ}\ $\uparrow$ & \textbf{BC}\ $\uparrow$ & \textbf{SR}\ $\uparrow$ & \textbf{Pick Score}\ $\uparrow$ & \textbf{FPS}\ $\uparrow$ & \textbf{Latency}\ $\downarrow$\\
\midrule
LiveEdit* &1.3B &-- &-- &-- &	1.542 &1.949 &1.640 &0.393 &19.26 &17.25 &\suboptimal{0.696}\\
SANA-Stream. &2B &0.379 &0.629 &0.381 &2.200 &1.777 &1.582 &0.428 &19.46 &\optimal{31.20} &0.772\\
Stream.V2 &14B &0.060 &0.885 &0.485 &1.316  &1.869 &0.473 &0.027 &\optimal{19.92} &\suboptimal{20.96} &\optimal{0.191}\\
\midrule
LucyEdit* &5B &-- &-- &-- &1.604  &1.797  &\optimal{2.121}  &0.326 &18.92 &3.083 &26.27\\
UniVideo &13B &\optimal{0.592} &0.605 &0.447 &2.377 &2.182 &1.866 &0.533 &19.54 &0.119 &680.9\\
Ditto &17B &0.334 &0.685 &0.355 &1.560  &1.796 &1.127 &0.167 &19.49 &0.294 &275.4\\
Qwen.+Wan. &17B &0.491 &\suboptimal{0.541} &\suboptimal{0.132} &\suboptimal{2.792} &\suboptimal{2.544} &1.897 &\suboptimal{0.713} &19.55 &1.274 &63.55\\
\midrule
Ours &17B &\suboptimal{0.550} &\optimal{0.499} &\optimal{0.124} &\optimal{2.796} &\optimal{2.609} &\suboptimal{2.024} &\optimal{0.720} &\suboptimal{19.61}
&14.47
&0.829\\
\bottomrule
\end{tabular}
}
\vspace{-1.1em}
\end{table*}

%% file: figs/ablation.tex
\begin{figure*}[t]
    \centering
    \includegraphics[width=\textwidth]{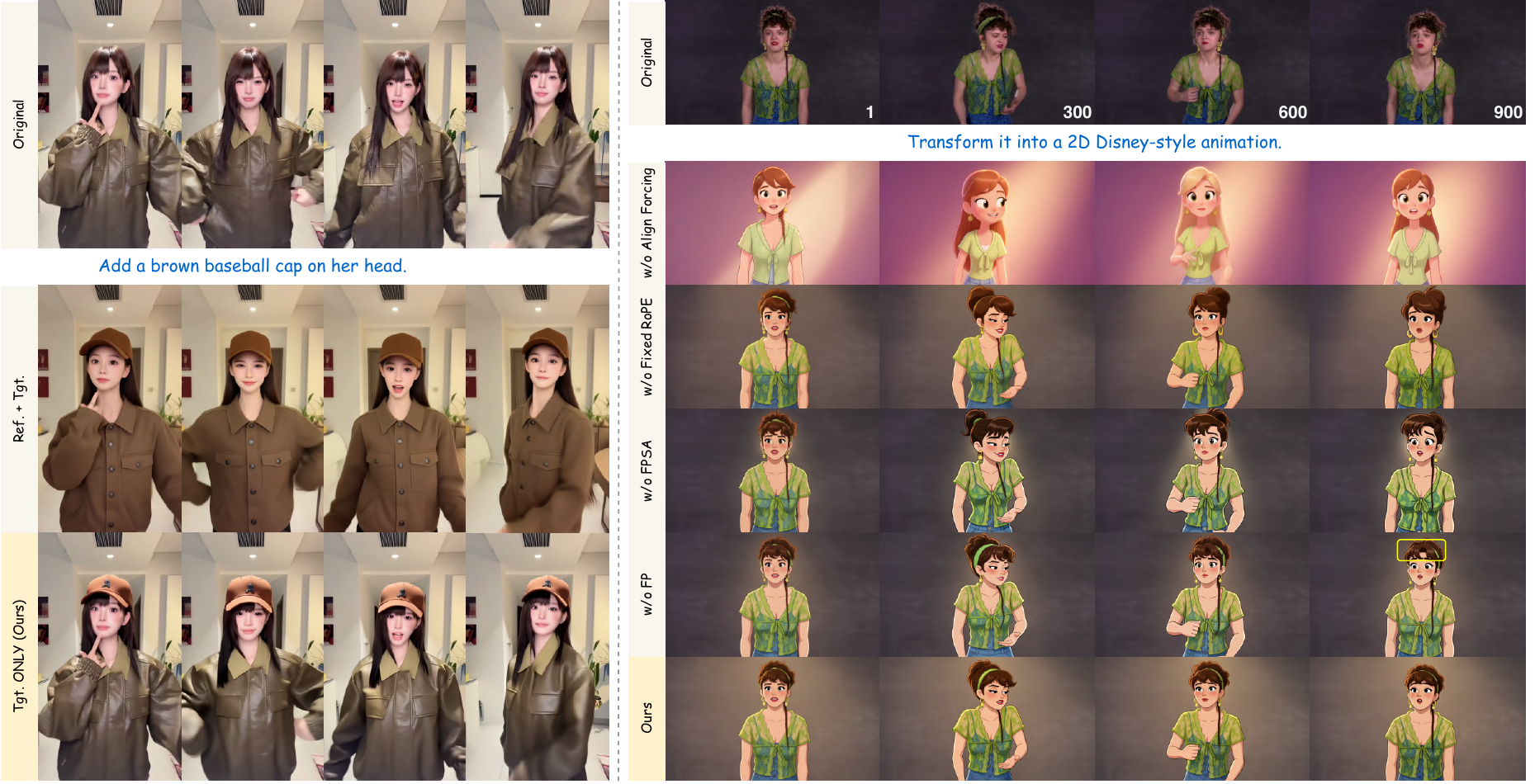}
    \vspace{-1.5em}
    \caption{Ablation study on causal streaming adaptation and aligned self-rollout distillation.
    }
    \label{fig:ablation}
    \vspace{-0.5em}
\end{figure*}

%% file: tables/ablation.tex
\begin{table*}[h]
\centering
\caption{Ablation study on aligned self-rollout distillation.}
\vspace{-1em}
\label{tab:quant_abl}
\resizebox{\textwidth}{!}{
\begin{tabular}{lccc|cccc|c|cc}
\toprule
\multirow{2}{*}{Method} & \multicolumn{3}{c}{\textbf{Character Consistency}} &\multicolumn{4}{c}{\textbf{VLM Evaluation}} &\multicolumn{1}{c}{\textbf{Video Quality}} &\multicolumn{2}{c}{\textbf{Temporal Consistency}}\\
\cmidrule(lr){2-4} \cmidrule(lr){5-8} \cmidrule(lr){9-9} \cmidrule(lr){10-11}
& \textbf{ID-SIM}\ $\uparrow$ & \textbf{AED}\ $\downarrow$ & \textbf{APD}\ $\downarrow$ & \textbf{TA}\ $\uparrow$ & \textbf{EQ}\ $\uparrow$ & \textbf{BC}\ $\uparrow$ & \textbf{SR}\ $\uparrow$ & \textbf{Pick Score}\ $\uparrow$ & \textbf{CLIP}\ $\uparrow$ & \textbf{DINO}\ $\uparrow$\\
\midrule
w/o Align Forcing &0.252 &0.711 &0.338 &2.703 &2.567 &1.306 &0.367 &\optimal{20.03} &\optimal{98.62} &\optimal{99.07}\\
w/o Fixed RoPE &0.381 &0.618 &0.120 &2.728 &2.639 &1.872 &0.617 &\suboptimal{19.91} &98.50 &98.95\\
w/o FPSA &0.455 &0.609 &0.116 &\suboptimal{2.737} &2.600 &2.105 &0.800 &19.85 &98.46	&98.88\\
w/o FP &\suboptimal{0.466} &\suboptimal{0.604} &\suboptimal{0.114} &2.725 &\suboptimal{2.622} &\suboptimal{2.139} &\optimal{0.833} &19.84	&98.42	&98.89\\
\midrule
Ours &\optimal{0.492} &\optimal{0.576} &\optimal{0.109} &\optimal{2.761} &\optimal{2.678} &\optimal{2.183} &\suboptimal{0.817} &19.78	&\suboptimal{98.51}	&\suboptimal{98.96}\\
\bottomrule
\end{tabular}
}
\vspace{-1em}
\end{table*}

%% file: sections/8_conclusion.tex
\section{Conclusion}
\label{sec:conclusion}

We introduce \SHORTNAME{}, a unified framework that brings instruction-guided character appearance editing to live video streams. By decoupling appearance from body motion and facial dynamics, our reconstruction-based training learns diverse appearance transformations from edited reference images and authentic, motion-aligned video targets; we construct CharEdit-50K to support this formulation. We further convert the bidirectional editing model into a chunk-wise causal generator and distill it into a two-step sampler. Align Forcing and Fixed RoPE align rollout training with streaming inference, while FPSA maintains stable appearance cues over long sequences with bounded historical context. Comprehensive experiments demonstrate the advantages of \SHORTNAME{} in terms of editing quality, motion consistency, long-term stability, and inference efficiency.

%% file: sections/appendix.tex
\newpage

\appendix

\section{Additional Information about Dataset}

\subsection{CharEdit-50K}
\label{app:data}

\noindent \textbf{Video Collection and Reference Selection.} To construct CharEdit-50K, we first assemble a diverse video pool from SpeakerVid-5M~\citep{speakervid}, the Seamless Interaction Dataset~\citep{seamless}, and additional videos collected from the Internet. From each video, we select a single high-quality frame as the appearance reference by examining facial sharpness and completeness, as well as the visibility and sharpness of the hands when present. Frames with severe occlusion, motion blur, truncation, or other visual degradations are excluded from selection, while each selected frame remains paired with its authentic source video. 

\noindent\textbf{Instruction Generation.} As shown in Fig.~\ref{fig:charedit}, conditioned on each reference image, GPT-5.5~\citep{gpt} generates candidate editing
instructions covering four categories. For \textit{add--remove}, the forward instruction adds an object or accessory to an unoccupied and physically plausible region, while the reverse instruction removes the added object. For \textit{remove--add}, the forward instruction removes an existing object and reconstructs the occluded content, while the reverse instruction adds the object with its visible attributes and placement. For \textit{change--restore}, the forward instruction changes only the color of a visible target, such as clothing or an accessory, while preserving its shape, material, and texture, and the reverse instruction specifies its original color. The final category is \textit{stylization}, which converts the reference image into a specified target visual style without requiring a corresponding sample in the reverse direction. After instruction generation, we balance the distributions of target categories and colors across the complete set of generated instruction pairs.

\noindent\textbf{Image Synthesis and Filtering.} Qwen-Image-Edit~\citep{qwen} and Nano Banana 2~\citep{gemini} synthesize candidate images according to the generated forward instructions. Given the original reference, an edited candidate, and its associated instruction(s), GPT-5.5 evaluates edit accuracy, target clarity, and character consistency. For bidirectional edits, it additionally evaluates whether the reverse instruction unambiguously specifies the inverse transformation.

\input{figs/charedit}

\input{figs/bench}

\subsection{CharEdit-Bench}
\label{app:bench}
To comprehensively evaluate character video editing in both conventional and streaming settings, we construct CharEdit-Bench with two subsets: CharEdit-Bench-S and CharEdit-Bench-L. CharEdit-Bench-S comprises 150 five-second videos, each containing 81 frames at a resolution of $480\times832$ and randomly paired with one editing prompt, yielding 150 evaluation cases. CharEdit-Bench-L comprises 30 videos at the same resolution, each longer than one minute at 16 FPS and paired with two editing prompts, resulting in 60 evaluation cases.  For every evaluation case, the first frame of the source video is used as the reference image. Examples from CharEdit-Bench are shown in Fig.~\ref{fig:bench}.

\section{Additional Implementation Details}

\input{tables/alignforcing}

\subsection{Training Details}
\label{app:implementation}

Our training pipeline proceeds through three stages. In Stage~1, we perform reconstruction-based appearance--motion decoupled training, enabling the model to learn instruction-guided appearance transformations while following independently provided motion conditions. We employ ViTPose~\citep{vitpose} to extract the skeleton sequence and localize the facial region, subsequently constructing the corresponding facial video. Following this initialization, Stage~2 adapts the bidirectional model to chunk-wise causal generation for streaming inference. We optimize only the LoRA adapters in the self-attention and FFN modules while keeping the cross-attention modules frozen~\citep{anyflow}. We implement the chunk-wise causal attention mask using PyTorch \texttt{flex\_attention}, allowing bidirectional attention within each chunk while restricting inter-chunk attention to the preceding clean context. After causal adaptation, Stage~3 performs aligned self-rollout distillation to compress the causal model into a two-step sampler and achieve long-term streaming generation, as shown in Algorithm~\ref{alg:aftrain}. We use the same trainable modules as in Stage~2. During DMD training, we optimize the student and fake-score branch while keeping the real-score branch frozen, performing five fake-score updates for every student update. To handle the high memory demand of Align Forcing training, we adopt Fully Sharded Data Parallel (FSDP) to reduce memory consumption. For FPSA, the query and key tokens are partitioned into spatio-temporal blocks of shape $(1,8,8)$. For style editing, synthesizing a complete stylized video for every training sample would be prohibitively expensive. We instead stylize a single image and repeat it along the temporal dimension to construct a static target video. Since appearance and motion are learned from decoupled conditions, this static target provides sufficient supervision for the target style, which can be combined with motion signals during inference.

\subsection{Baseline Configurations}
\label{app:baseline}

All baselines are evaluated using their official implementations and released checkpoints. The \#Params column in Table~\ref{tab:quant_rec} reports only the number of parameters in the DiT backbone used by each method. For Ditto~\citep{ditto}, we use the officially released \texttt{ditto\_local.safetensors} and \texttt{ditto\_global\_style.safetensors} LoRA checkpoints for local and global style editing, respectively. Since SANA-Streaming~\citep{sana-streaming} was trained at a resolution of $704\times1280$, we resize the input videos to this resolution for inference and resize the generated videos back to $480\times832$ before evaluation. For StreamDiffusionV2~\citep{streamdiffusionv2}, we follow its official inference launcher, which uses one-step denoising by default.

\subsection{Evaluation Details}
\label{app:evaluation}
We evaluate identity and motion consistency between the source and edited videos using three paired metrics. ID-SIM is computed as the average cosine similarity between ArcFace~\citep{arcface} embeddings extracted from temporally corresponding source and edited frames. AED~\citep{fomm} is calculated as the average $\ell_1$ distance between extracted expression parameters of the source and edited frames using SMIRK~\citep{smirk}. APD~\citep{fomm} is computed as the average $\ell_1$ distance between corresponding body keypoints extracted by ViTPose~\citep{vitpose}. Higher ID-SIM and lower AED and APD indicate better preservation of identity, facial expressions, and body poses, respectively. For VLM evaluation, we uniformly sample three temporally aligned frame pairs from each source and edited video and provide GPT-5.5~\citep{gpt} with the source frame, edited frame, and editing instruction. GPT-5.5 assigns a score from $0$ to $3$ for each of three criteria: Text Alignment, Edit Quality, and Background Consistency. It additionally returns a binary overall-success judgment based on these criteria. Based on these binary judgments, we define edit Success Rate as the proportion of sampled source--edited frame pairs judged successful. We use Pick Score~\citep{pickscore} to assess overall video quality. For inference efficiency, we report FPS and inter-chunk latency on a single NVIDIA H100 GPU using 81-frame clips at a resolution of $384\times672$~\citep{helios,liveavatar}. Runtime is measured over the generation pipeline, from model-ready conditions to decoded RGB frames. For each model, we enable its officially supported acceleration techniques (e.g., FlashAttention, \texttt{torch.compile}, and warm-up) to maximize throughput. Specifically, \SHORTNAME{} is compiled with \texttt{torch.compile}, and its reported runtime includes motion-condition encoding, two-step DiT denoising, and VAE decoding. For long-video evaluation, we assess temporal consistency using frame-wise CLIP~\citep{clip} and DINO~\citep{dino} feature similarities.

\section{More Experimental Results}
\label{more_experiments}

\noindent \textbf{Evaluation on Long Videos.} We further evaluate long-term character video editing in the streaming setting on CharEdit-Bench-L. As shown in Table~\ref{tab:quant_long}, \SHORTNAME{} achieves the strongest overall performance among the compared methods, demonstrating that its editing quality and character consistency are maintained over long sequences. The qualitative comparisons in Fig.~\ref{fig:qualitative_long} further show that \SHORTNAME{} maintains a consistent edited appearance over long video sequences while faithfully following the source motion. These results demonstrate the long-term stability of \SHORTNAME{}.

\input{figs/qualitative_long}
\input{tables/comparison_long}

\noindent \textbf{Cross-Character Editing.} Owing to our appearance--motion decoupled formulation, \SHORTNAME{} naturally supports cross-character editing by combining the appearance of a reference character with motion signals extracted from a different driving video, as shown in Fig.~\ref{fig:crossid}.

\input{figs/crossid}

\noindent \textbf{Additional Qualitative Results.} Figs.~\ref{fig:qualitative_1}, \ref{fig:qualitative_2}, \ref{fig:qualitative_3}, and~\ref{fig:qualitative_4} present additional qualitative comparisons across diverse characters and editing instructions, further demonstrating the robustness and generalization of \SHORTNAME{}. Figs.~\ref{fig:long_video_1} and~\ref{fig:long_video_2} present comparisons over extended video sequences, highlighting the long-term consistency and stability of \SHORTNAME{}. Compared with existing methods, \SHORTNAME{} exhibits substantially less appearance drift and consistently maintains the requested edit and character appearance while following the source motion.

\section{Limitations \& Future Work}
While \SHORTNAME{} achieves real-time and temporally coherent long-term character video editing, there remain two directions for further improvement, as shown in Fig.~\ref{fig:limitation}. First, the current framework relies on skeleton sequences for body-motion control. Although they effectively capture overall body poses, they do not explicitly represent fine-grained finger articulation, and subtle or complex hand gestures may therefore be reproduced less accurately. Incorporating more expressive hand-motion conditions, such as dense hand keypoints or hand-specific representations, could further improve fine-grained motion preservation. Second, character appearance is represented by a single reference image. Although this compact condition is effective in most cases, it provides limited appearance cues for regions occluded in the reference image. When these regions become visible during subsequent motion, their inferred appearance may differ from that in the source video. Future work could leverage multiple reference images to provide more complete appearance cues.

\input{figs/limitation}

\section{Ethics Statement}
Our work focuses on enabling real-time and long-term character video editing, which naturally raises concerns related to privacy, consent, copyright, and potential misuse. We respect the applicable licenses and usage policies of the data used for training and evaluation. The data are obtained from existing research datasets and publicly accessible online sources, and we do not redistribute third-party source videos or audio. Our method is not designed to identify individuals or recover private identity information. While realistic character editing may be susceptible to impersonation, deceptive manipulation, or non-consensual use, EditaLive is intended solely for legitimate creative production and interactive applications. We encourage users to obtain consent from depicted individuals and adopt responsible deployment practices, such as access control, content disclosure, and watermarking, to mitigate malicious use.

\input{figs/more_results}

%% file: figs/charedit.tex
\begin{figure*}[ht]
    \centering
    \includegraphics[width=\textwidth]{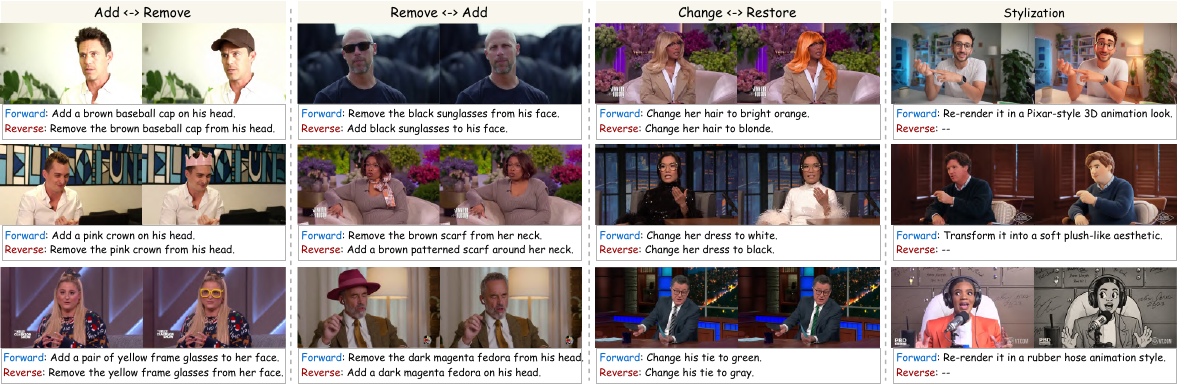}
    \vspace{-1.5em}
    \caption{Examples from CharEdit-50K.}
    \label{fig:charedit}
    \vspace{-0.5em}
\end{figure*}

%% file: figs/bench.tex
\begin{figure*}[ht]
    \centering
    \includegraphics[width=\textwidth]{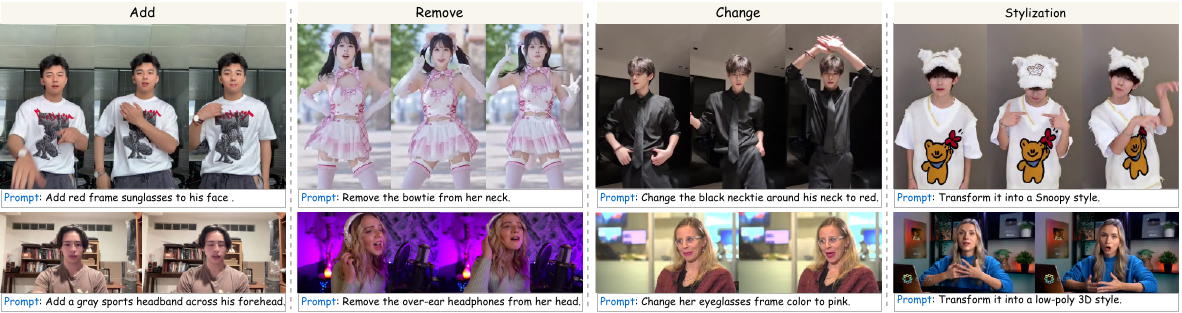}
    \vspace{-1.5em}
    \caption{Examples from CharEdit-Bench.}
    \label{fig:bench}
    \vspace{-1em}
\end{figure*}

%% file: tables/alignforcing.tex
\begin{algorithm}[!t]
\setlength{\baselineskip}{1.1\baselineskip}
\caption{Align Forcing Training with Fixed RoPE}
\label{alg:aftrain}
\begin{algorithmic}[1]
\Require Denoising timesteps $\{t_0,t_1, \dots, t_T\}$, where $t_0 = 0$
\Require Number of video chunks $N$
\Require Reference frame $x_R$
\Require Per-chunk conditions $\{C_i\}_{i=1}^{N}$, including text and motion signals
\Require Cache lengths $(L_{\mathrm{ref}},L_{\mathrm{sink}},L_{\mathrm{local}})=(1,3,3)$
\Require Generator $G_{\theta}$, which also returns pre-RoPE KV features
\Loop
    \State Initialize model output $\mathbf{X}_{\theta} \gets []$
    \State Initialize $\KVSet_\mathrm{sink} \gets []$, $\KVSet_\mathrm{local} \gets []$
    \State Sample $s \sim \text{Uniform}\{1, 2, \ldots, T\}$
    \State Set reference KV cache $\KVSet_\mathrm{ref} \gets sg(G_\theta(x_R;t_0))$
    \For{$i = 1, \dots, N$}
        \State Initialize $x^i_{t_T} \sim \mathcal{N}(0,I)$
        \State $\KVSet^{i} \gets [\KVSet_\mathrm{ref}, \KVSet_\mathrm{sink}, Tail_{L_\mathrm{local}}(\KVSet_\mathrm{local})]$ \Comment{fixed-length active cache}
        \State Assign Fixed RoPE indices $(0{:}9)$
        \For{$j = T, \dots, 1$} \Comment{complete the full rollout}
            \If{$j = s$}
                \State Enable gradient computation
                \State Set $\KVOutput^i$, $\hat{x}^i_{0} \gets G_\theta(x^i_{t_{j}}; t_j, \KVSet^i,C_i)$
                \State $\ModelOutput{\texttt{.append}}(\hat x^i_{0})$
                \State Detach $\KVOutput^i$ from gradient graph
            \Else
                \State Disable gradient computation
                \State Set $\KVOutput^i$, $\hat{x}^i_{0} \gets G_\theta(x^i_{t_j}; t_j, \KVSet^i,C_i)$
            \EndIf
            \If{$j>1$}
                \State Sample $\epsilon \sim \mathcal{N}(0, I)$
                \State Set $x^i_{t_{j-1}} \gets \Psi(\hat{x}^i_0, \epsilon, t_{j-1})$
            \EndIf
        \EndFor
        \If{$i = 1$}
            \State Disable gradient computation
            \State Set $\KVOutput^i$, $\_ \gets G_\theta(\hat{x}^i_{0}; t_0, \KVSet^i,C_i)$ \Comment{one-time cache forward on the first chunk}
            \State $\KVSet_\mathrm{sink}{\texttt{.append}}(\KVOutput^i)$
        \Else
            \State $\KVSet_\mathrm{local}{\texttt{.append}}(\KVOutput^i)$
        \EndIf
    \EndFor
    \State Update $\theta$ via DMD loss
\EndLoop
\end{algorithmic}
\end{algorithm}

%% file: figs/qualitative_long.tex
\begin{figure*}[t]
    \centering
    \includegraphics[width=\textwidth]{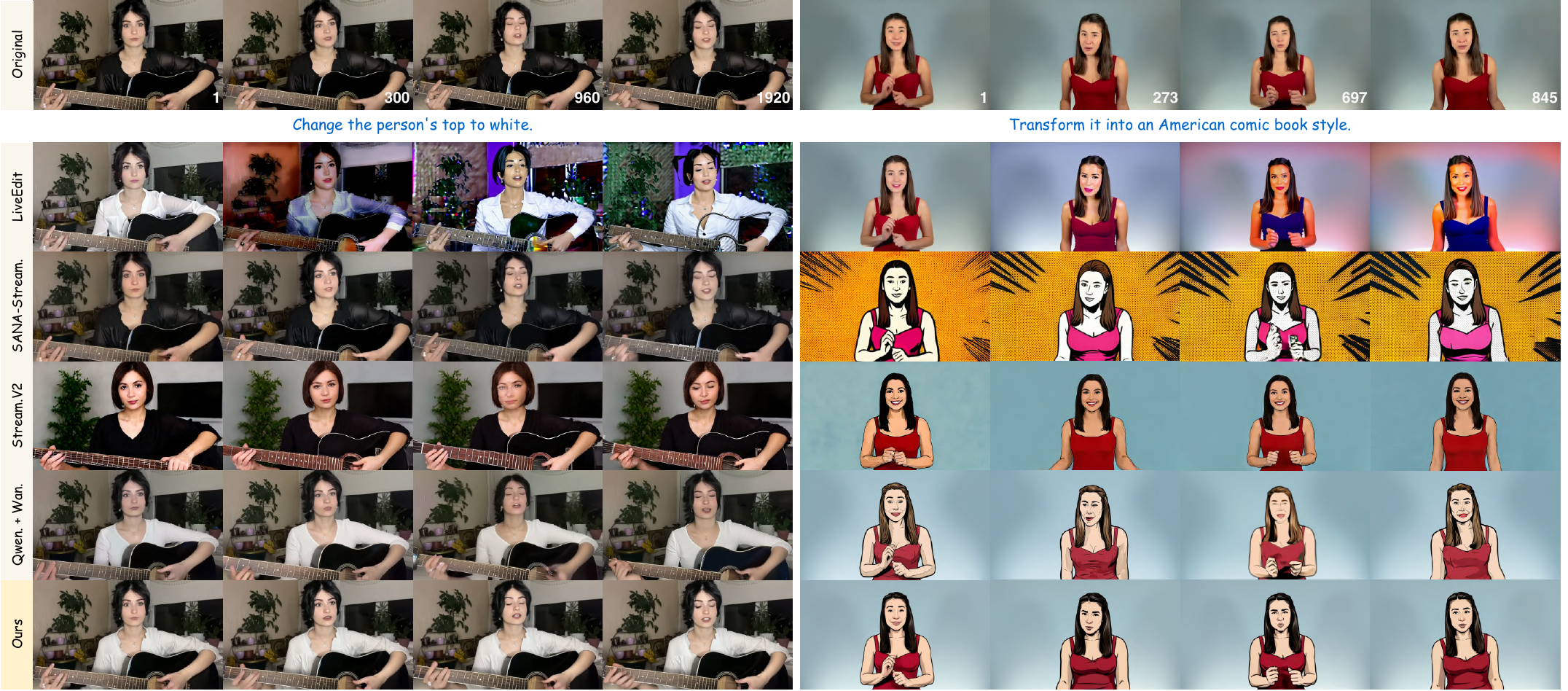}
    \vspace{-1.5em}
    \caption{Qualitative comparisons on CharEdit-Bench-L. \SHORTNAME{} consistently maintains the requested edit and character appearance while following the source motion.
    }
    \label{fig:qualitative_long}
    % \vspace{-1em}
\end{figure*}

%% file: tables/comparison_long.tex
\begin{table*}[t]
% \vspace{-1em}
\centering
\caption{Quantitative comparisons on CharEdit-Bench-L. Numbers in \textcolor{myred}{\textbf{red}} and \textcolor{myblue}{\textbf{blue}} indicate the best and the second-best results, respectively. APD multiplied by $10$. \textit{*LiveEdit does not support global style editing. We therefore omit its character-consistency metrics, as these scores are not comparable when the requested edit is not successfully performed.}}
\label{tab:quant_long}
\resizebox{\textwidth}{!}{
\begin{tabular}{lccc|cccc|c|cc}
\toprule
\multirow{2}{*}{Method} & \multicolumn{3}{c}{\textbf{Character Consistency}} &\multicolumn{4}{c}{\textbf{VLM Evaluation}} &\multicolumn{1}{c}{\textbf{Video Quality}} &\multicolumn{2}{c}{\textbf{Temporal Consistency}}\\
\cmidrule(lr){2-4} \cmidrule(lr){5-8} \cmidrule(lr){9-9} \cmidrule(lr){10-11}
& \textbf{ID-SIM}\ $\uparrow$ & \textbf{AED}\ $\downarrow$ & \textbf{APD}\ $\downarrow$ & \textbf{TA}\ $\uparrow$ & \textbf{EQ}\ $\uparrow$ & \textbf{BC}\ $\uparrow$ & \textbf{SR}\ $\uparrow$ & \textbf{Pick Score}\ $\uparrow$ & \textbf{CLIP}\ $\uparrow$ & \textbf{DINO}\ $\uparrow$\\
\midrule
% StreamV2V \\
% \midrule
LiveEdit* &-- &-- &-- &1.583 &1.700 &1.044 &0.017 &18.91 &97.36 &98.38\\
SANA-Stream. &0.233	&0.725	&0.402 &2.272 &1.938 &1.121 &0.233 &19.62 &97.28	&98.25\\
Stream.V2 &0.049 &0.876 &0.329 &1.431 &2.064 &0.638 &0.100 & \suboptimal{19.70} &\suboptimal{98.42} &\optimal{99.17}\\
Qwen.+Wan. &\suboptimal{0.398} &\suboptimal{0.659} &\suboptimal{0.126} &\suboptimal{2.733} &\suboptimal{2.304} &\suboptimal{1.776} &\suboptimal{0.750} &19.69 &98.30 &98.65\\
\midrule
Ours &\optimal{0.492} &\optimal{0.576} &\optimal{0.109} &\optimal{2.761} &\optimal{2.678} &\optimal{2.183} &\optimal{0.817} &\optimal{19.78}	&\optimal{98.51}	&\suboptimal{98.96}\\
\bottomrule
\end{tabular}
}
\vspace{-1em}
\end{table*}

%% file: figs/crossid.tex
\begin{figure*}[t]
    \centering
    \includegraphics[width=\textwidth]{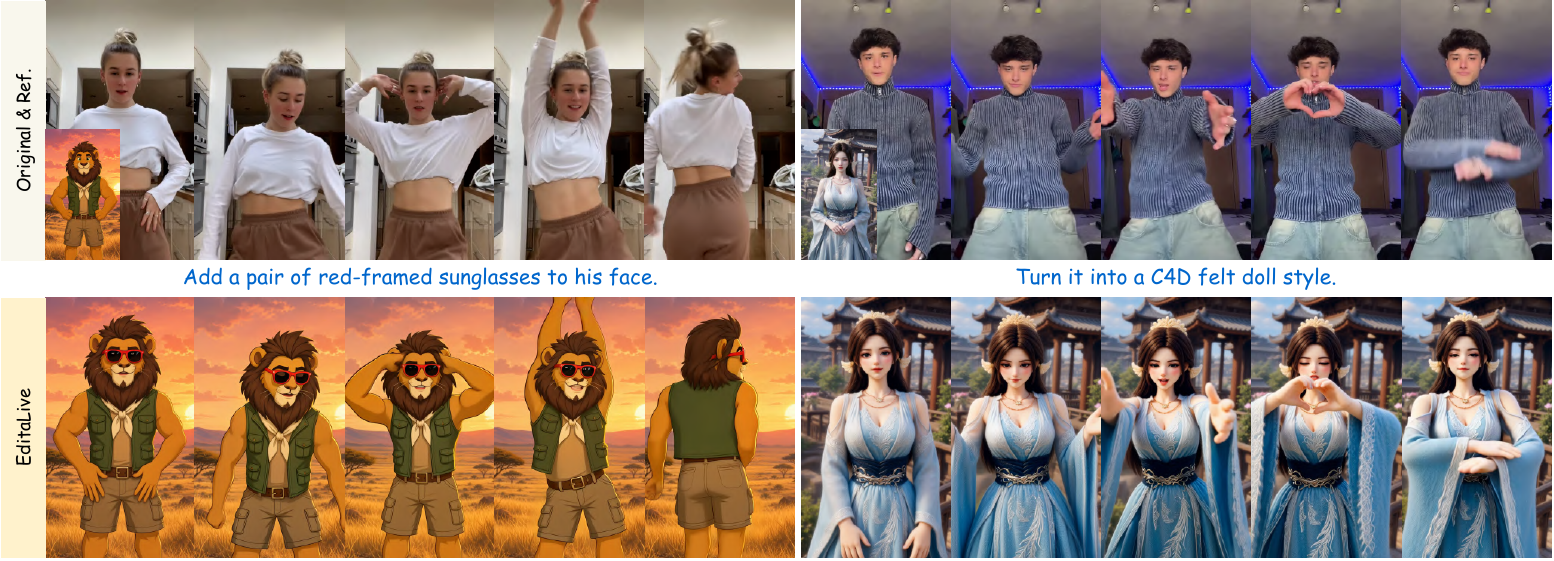}
    \vspace{-1.5em}
    \caption{Cross-character editing results.
    }
    \label{fig:crossid}
    % \vspace{-1em}
\end{figure*}

%% file: figs/limitation.tex
\begin{figure*}[t]
    \centering
    \includegraphics[width=\textwidth]{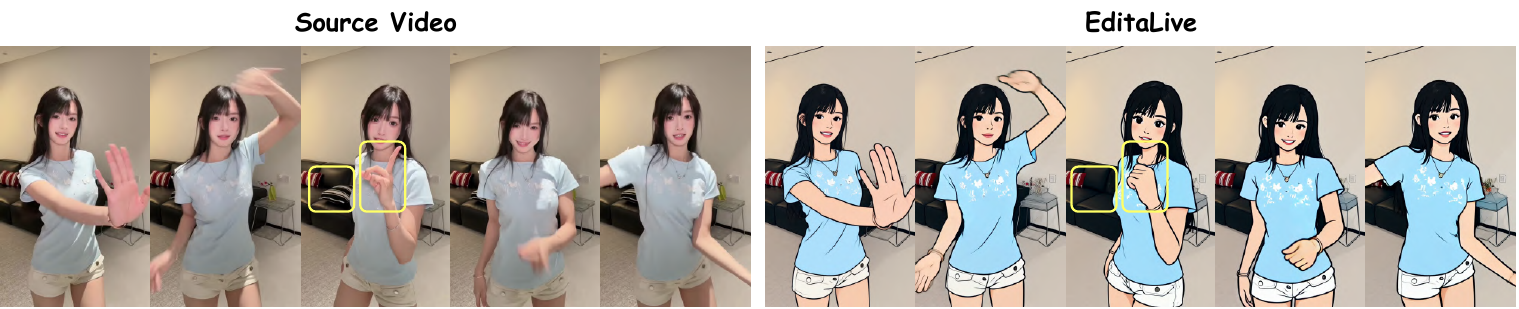}
    \vspace{-1.5em}
    \caption{Limitations of \SHORTNAME. The editing prompt is ``\textit{Transform it into a Snoopy style}''.
    }
    \label{fig:limitation}
    % \vspace{-1em}
\end{figure*}

%% file: figs/more_results.tex
\begin{figure*}[t]
    \centering
    \includegraphics[width=0.9\textwidth]{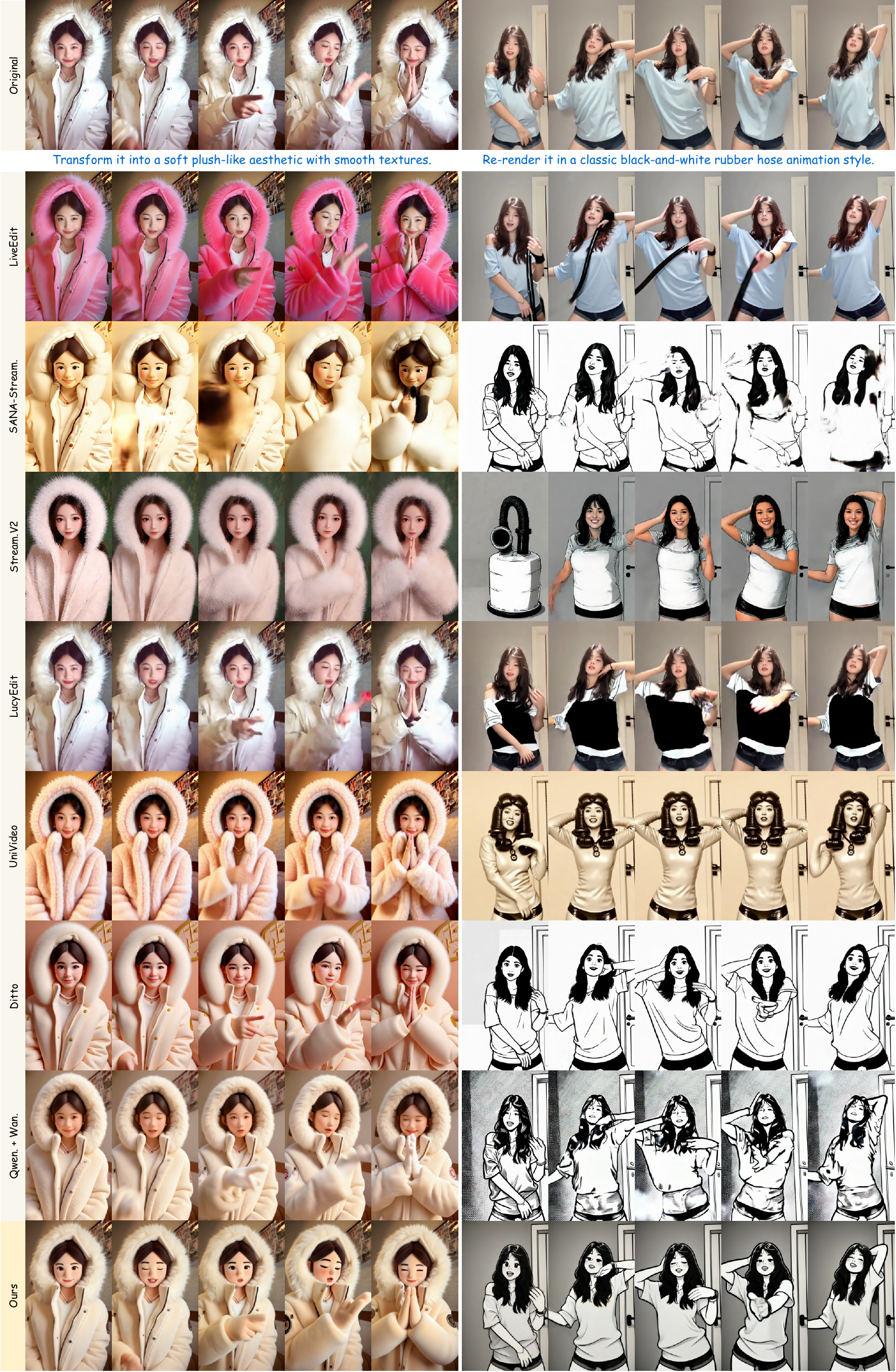}
    % \vspace{-2em}
    \caption{More comparisons between baselines and our \SHORTNAME{} (1/4).
    }
    \label{fig:qualitative_1}
    % \vspace{-1em}
\end{figure*}

\begin{figure*}[t]
    \centering
    \includegraphics[width=0.9\textwidth]{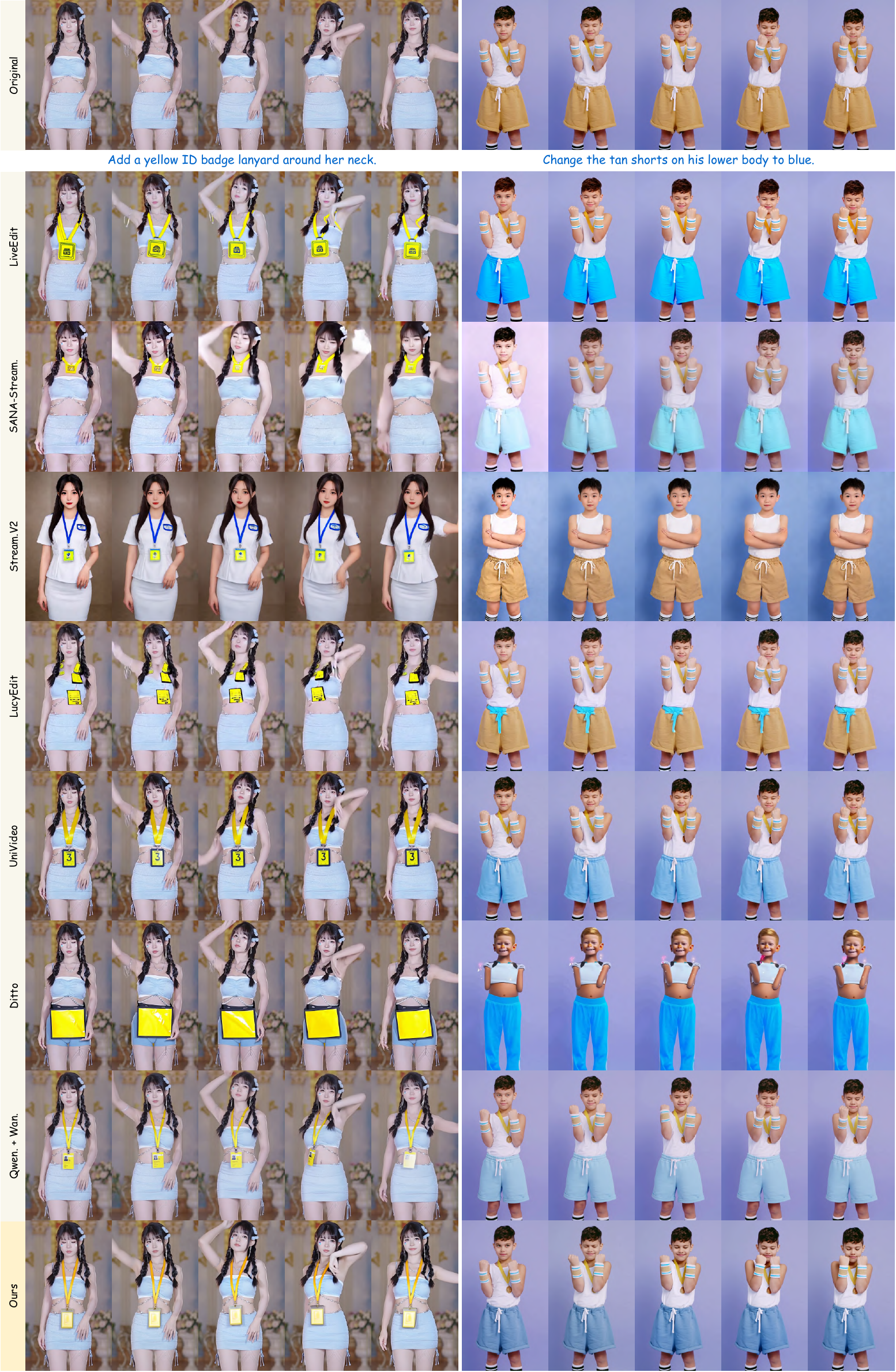}
    % \vspace{-2em}
    \caption{More comparisons between baselines and our \SHORTNAME{} (2/4).
    }
    \label{fig:qualitative_2}
    % \vspace{-1em}
\end{figure*}

\begin{figure*}[t]
    \centering
    \includegraphics[width=\textwidth]{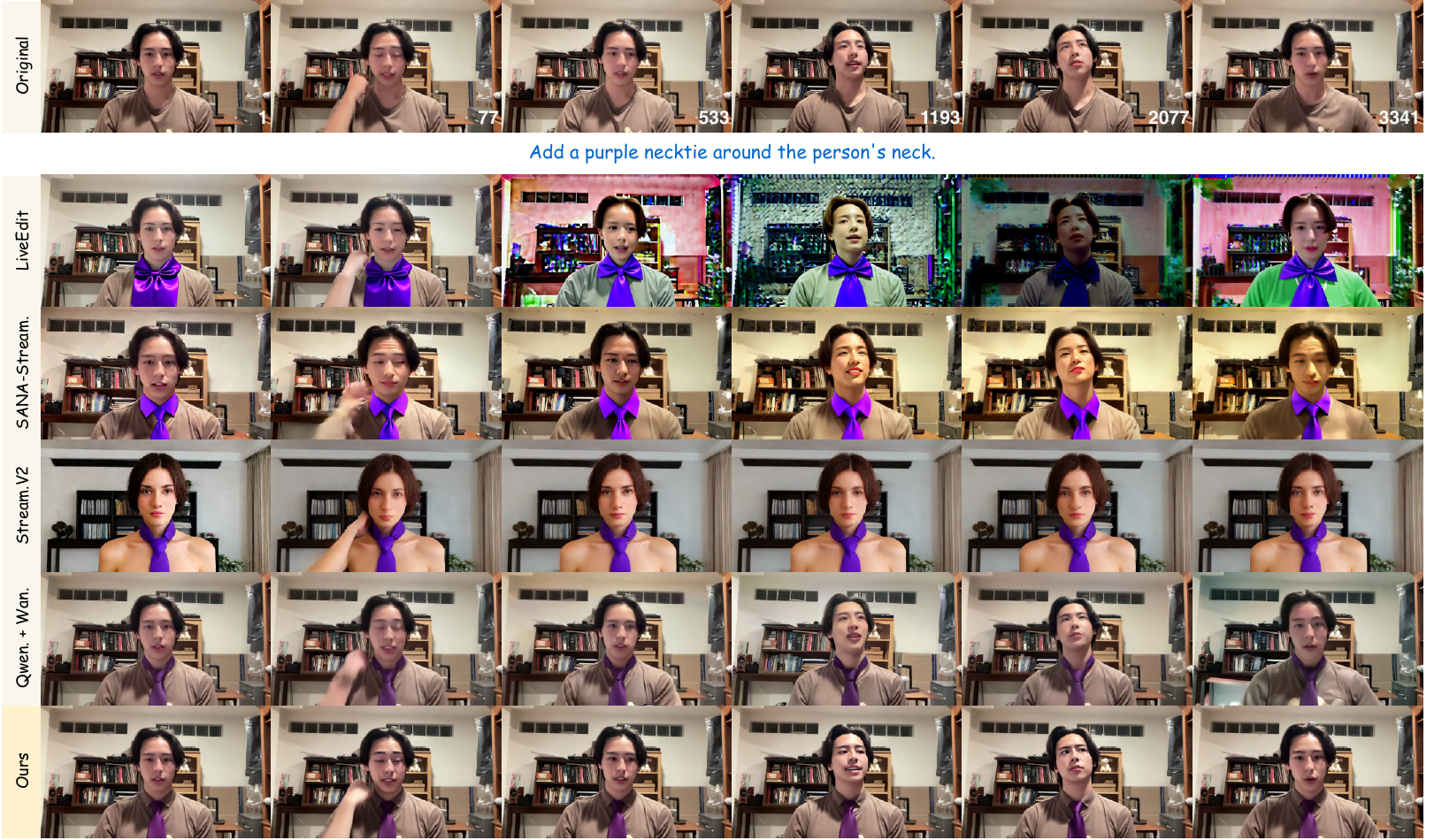}
    \vspace{-1em}
    \caption{Long-video comparison results (1/2).
    }
    \label{fig:long_video_1}
\end{figure*}

\begin{figure*}[t]
    \centering
    \includegraphics[width=\textwidth]{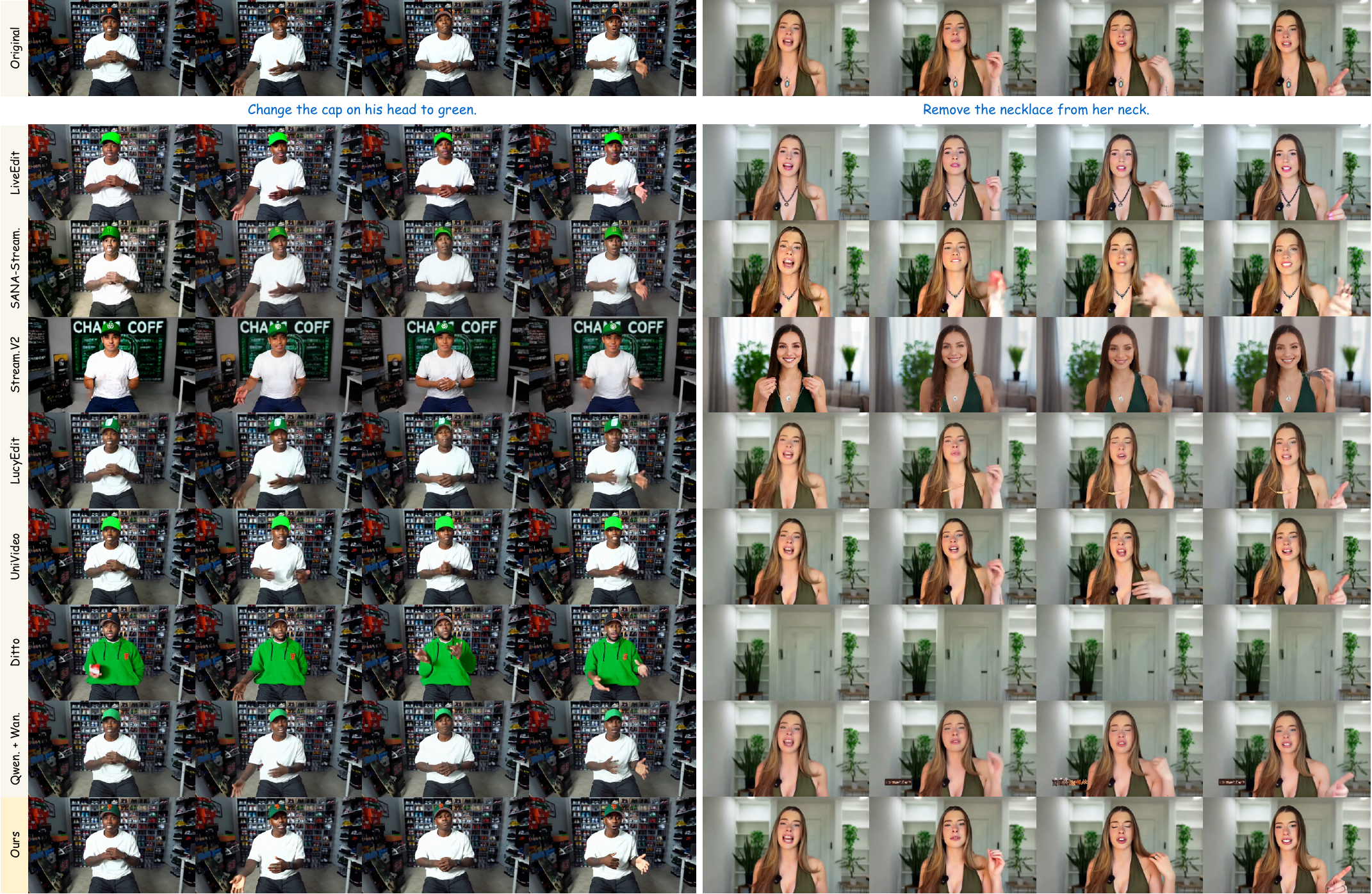}
    \vspace{-1em}
    \caption{More comparisons between baselines and our \SHORTNAME{} (3/4).
    }
    \label{fig:qualitative_3}
    % \vspace{-1em}
\end{figure*}

\begin{figure*}[t]
    \centering
    \includegraphics[width=\textwidth]{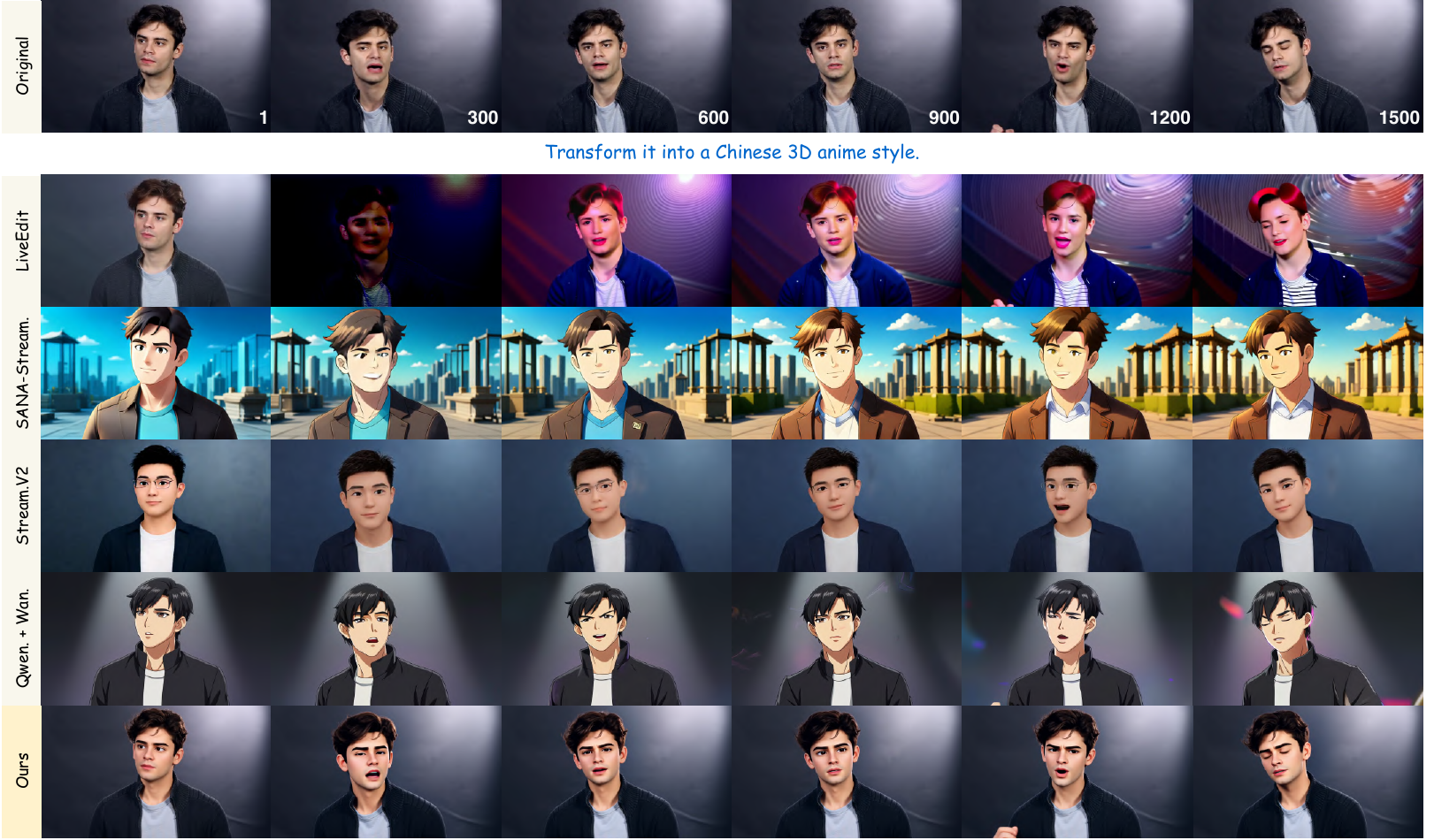}
    \vspace{-1em}
    \caption{Long-video comparison results (2/2).
    }
    \label{fig:long_video_2}
\end{figure*}

\begin{figure*}[t]
    \centering
    \includegraphics[width=\textwidth]{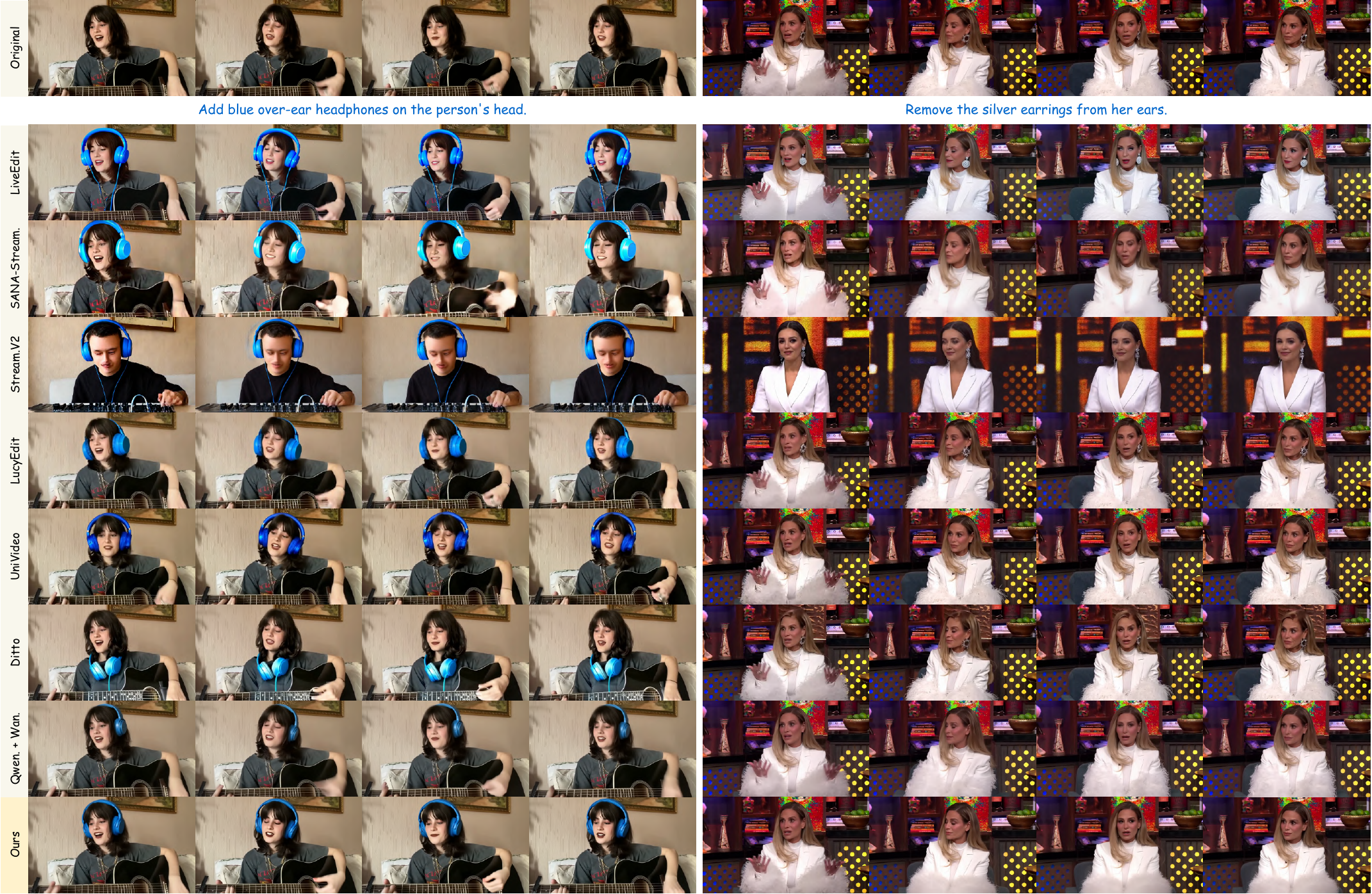}
    \vspace{-1em}
    \caption{More comparisons between baselines and our \SHORTNAME{} (4/4).
    }
    \label{fig:qualitative_4}
    % \vspace{-1em}
\end{figure*}